\documentclass[10pt,twocolumn,letterpaper]{article}

\usepackage[pagenumbers]{wacv} 

\usepackage[accsupp]{axessibility}  
\usepackage{graphicx}
\usepackage{amsmath}
\usepackage{amssymb}
\usepackage{booktabs}
\usepackage{xcolor}

\usepackage[pagebackref,breaklinks,colorlinks]{hyperref}

\usepackage[capitalize]{cleveref}
\crefname{section}{Sec.}{Secs.}
\Crefname{section}{Section}{Sections}
\Crefname{table}{Table}{Tables}
\crefname{table}{Tab.}{Tabs.}

\def\wacvPaperID{247} 
\def\confName{WACV}
\def\confYear{2024}

\begin{document}

\title{HALSIE: \underline{H}ybrid \underline{A}pproach to \underline{L}earning \underline{S}egmentation by Simultaneously Exploiting \underline{I}mage and \underline{E}vent Modalities}
\author{Shristi Das Biswas, Adarsh Kosta, Chamika Liyanagedera, Marco Apolinario and Kaushik Roy\\
Purdue University, West Lafayette, Indiana, USA\\
{\tt\small \{sdasbisw, akosta, cliyanag, mapolina, kaushik\}@purdue.edu}}
\maketitle

\begin{abstract}
Event cameras detect changes in per-pixel intensity to generate asynchronous `event streams'. They offer great potential for accurate semantic map retrieval in real-time autonomous systems owing to their much higher temporal resolution and high dynamic
range (HDR) compared to
conventional cameras. 
However, existing implementations for event-based segmentation suffer from sub-optimal performance since these temporally dense events only measure the varying component of a visual signal, limiting their ability to encode dense spatial context compared to frames. To address this issue, we propose a hybrid end-to-end learning framework HALSIE, utilizing three key concepts to reduce inference cost by up to $20\times$ versus prior art while retaining similar performance: 
First, a simple and efficient cross-domain learning scheme to extract complementary spatio-temporal embeddings from both frames and events. 
Second, a specially designed dual-encoder scheme with Spiking Neural Network (SNN) and Artificial Neural Network (ANN) branches
to minimize latency while retaining cross-domain feature aggregation. Third, a multi-scale cue mixer to model rich representations of the fused embeddings. These qualities of HALSIE allow for a very lightweight architecture 
achieving state-of-the-art segmentation performance on DDD-17, MVSEC, and DSEC-Semantic datasets with up to $33\times$ higher parameter efficiency and 
favorable inference cost (17.9mJ per cycle). 
Our ablation study also brings new insights into effective design choices that can prove beneficial for research across other vision tasks.
\end{abstract}

\vspace{-3mm}
\section{Introduction}
\label{sec:intro}


\begin{figure}
  \centering
  \includegraphics[width=1.0\linewidth, height = 0.6\linewidth]{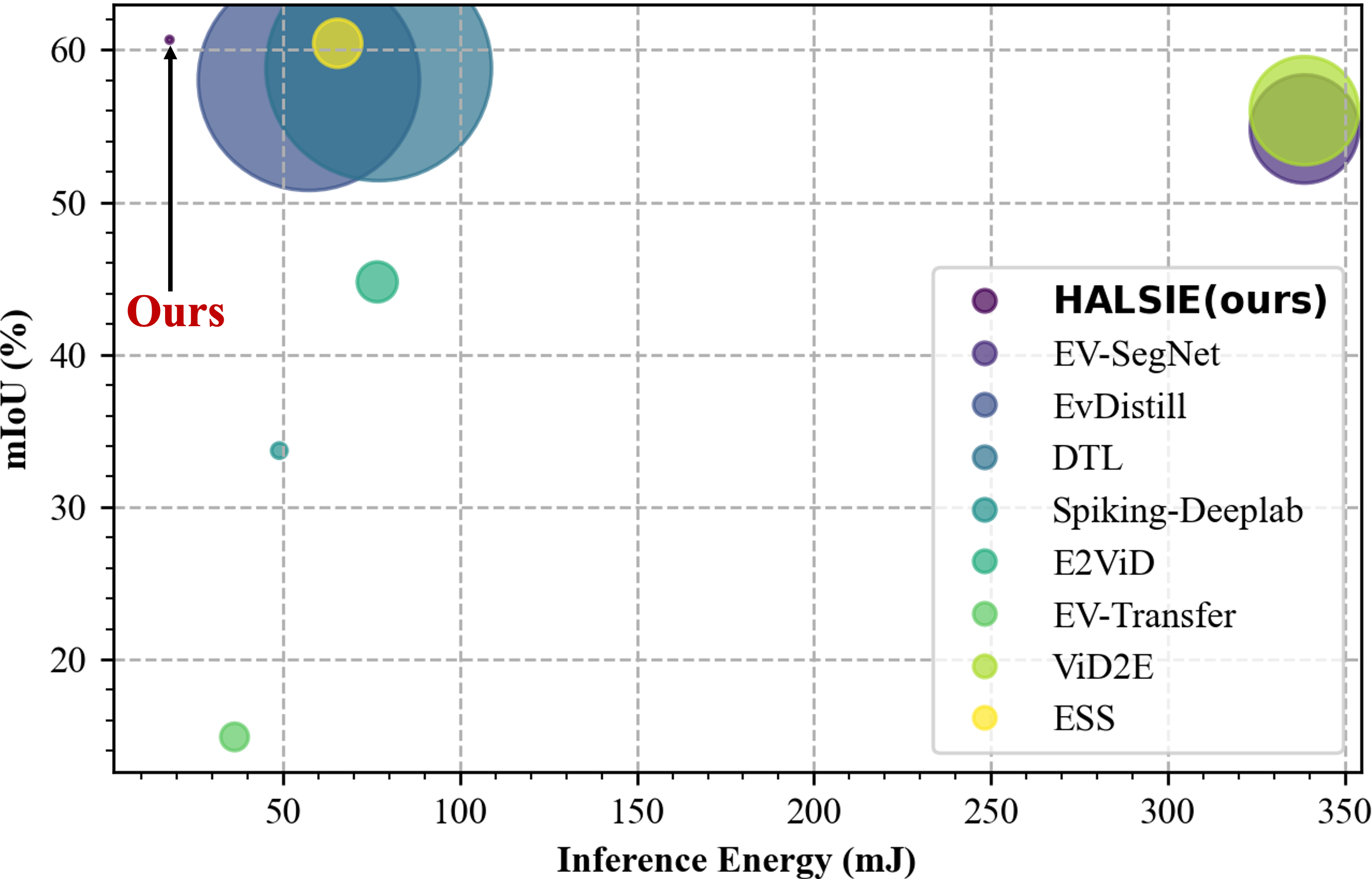}\vspace{-2mm}
  \caption{\textbf{Segmentation performance vs inference energy} of our HALSIE model on DDD-17 dataset evaluated on 45nm CMOS process. The circle areas are proportional to the model size.}
  \vspace{-3mm}
  \label{fig:plot}
\end{figure}


We often see house-flies seamlessly navigating through cluttered spaces, supporting such complex movements with just a few million neurons~\cite{borst2010fly}. 
Yet, modern autonomous systems with significantly higher compute capability and near unlimited resources~\cite{sze2017efficient} still fail to replicate the comprehensive real-time scene understanding achieved by these tiny biological systems with very low power budgets. In this work, we discuss these shortcomings for the task of semantic segmentation which is an important building block in the perception pipeline for 
autonomous navigation systems.

Present-day segmentation falls back on conventional frame cameras that share little with the biological eye in their scene capturing mechanism~\cite{honeybee1996, baird2013universal}. They sample intensity frames synchronously at constant and large time intervals. In addition, they fail to capture information in challenging scenarios with HDR and motion blur, leading to the loss of essential scene details. In safety-critical applications like automotive, this may come at the cost of fatalities. Increasing the sampling rate would enable them to capture high speed motion but leads to redundant background information retrieval along with elevated energy consumption.

\begin{figure}
  \centering
  \begin{subfigure}{0.5\linewidth}
    \includegraphics[width=\linewidth]{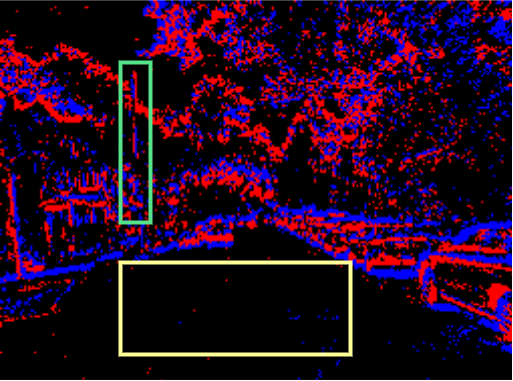}
    \caption{} 
    \label{fig:1a}
  \end{subfigure}%
  \hspace*{\fill}   
  \begin{subfigure}{0.5\linewidth}
    \includegraphics[width=\linewidth]{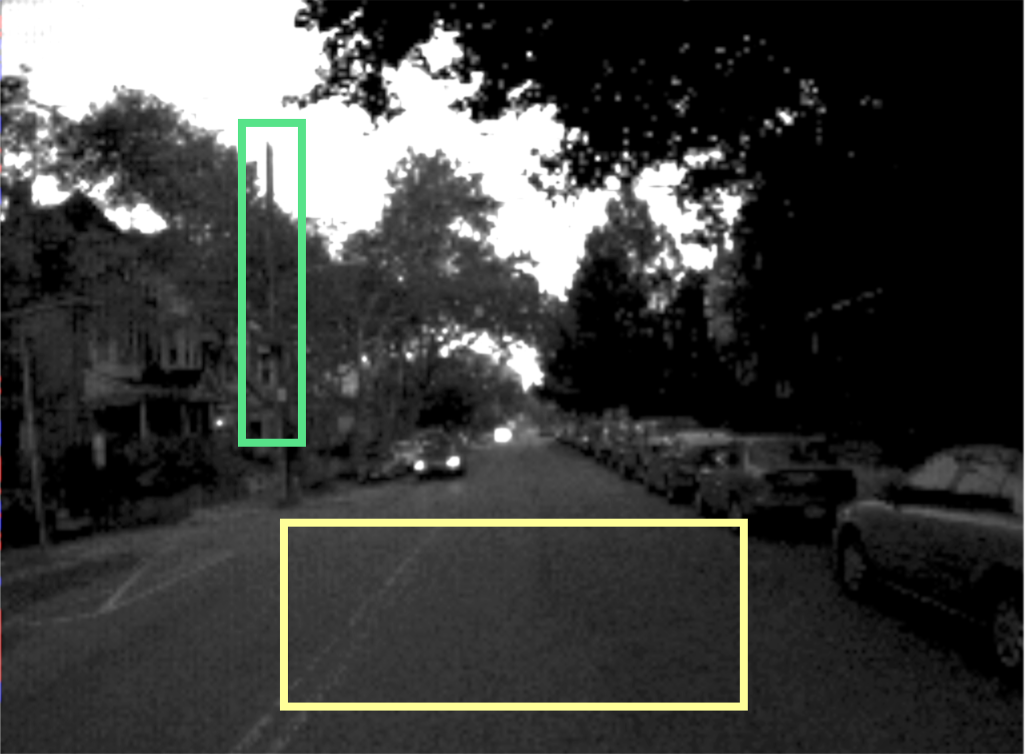}
    \caption{} 
    \label{fig:1b}
  \end{subfigure}%
  \vspace{-3mm}
  \caption{\textbf{Complementary feature extraction.} (a) Events capture temporally rich features only where intensity changes occur, (b) lacking the dense spatial information contained in standard frames.}
  \vspace{-5mm}
  \label{fig:eventsparse}
\end{figure}

To circumvent these issues, researchers have explored event cameras~\cite{lichtsteiner, brandli2014240, delbruck2010activity} as an alternate sensing modality. Event sensors asynchronously measure changes in per-pixel intensity to output sparse data streams at high temporal resolution (10µs vs 3ms), higher dynamic range (140dB vs 60dB) and significantly lower energy (10mW vs 3W) compared to frame cameras. These properties make event cameras enticing for high-speed segmentation. However, the event stream only contains information about pixels experiencing intensity changes, rendering the retrieval of dense contextual information from scenes challenging~\cite{lee2020spike}. As seen in Fig.~\ref{fig:eventsparse}, intensity information is absent at pixels where events are not recorded leading to scenes with incomplete contextual information (highlighted in a yellow box). On the other hand, the electric pole (green box) is concealed in the frame output due to high dynamic range. Both cameras are, therefore, complementary. This complementarity motivates the development of a multi-modal hybrid approach to improve
semantic performance in challenging conditions, leveraging advantages of both the frame- and event-domains.

Traditional learning methods based on Artificial Neural Networks (ANNs), originally
designed for frame-based images~\cite{tao2020hierarchical, yuan2019segmentation, yuan2021ocnet, zhao2017pyramid, chen2018encoder}, are inept at directly handling discrete and asynchronous event streams~\cite{gallego2020event}. Several works propose to convert asynchronous events to standard frames for downstream processing based on handcrafted features~\cite{lagorce2016hots, maqueda2018event, rebecq2017real}. Unfortunately, unlike frame-camera images, such accumulated event frames are much sparser and lack dense semantic information. To further aggravate the
issue, ANNs discard the temporal ordering of these inputs by representing them as parallel channels and perform sub-optimal stateless computations on the sequential data stream. Recent work has shown that such stateless
ANNs show inferior performance on event-based semantic segmentation, only improving when they leverage temporal recurrence~\cite{sun2022ess, rebecq2019high}. Nevertheless, these methods require large model sizes and incur high inference costs ($>50$ mJ per cycle) since they do not rely on efficient processing to exploit the sparse information retrieval aspect of event cameras (Fig.~\ref{fig:plot}). 


Recently, Spiking Neural Networks (SNN)~\cite{roy2019towards} have emerged as promising candidates for directly handling event streams. They perform asynchronous sparse event-driven compute~\cite{wang2022hierarchical} and offer an implicit recurrence through their internal states to preserve input temporal information~\cite{ponghiran2022spiking} in a stateful manner. Such traits make SNNs ideal for handling sequential inputs efficiently, enabling considerable energy savings on neuromorphic hardware~\cite{davies2018loihi, akopyan2015truenorth, furber2014spinnaker}.

With this in mind, we fundamentally revisit the design of end-to-end-learning frameworks for semantic segmentation using events and frames. In particular, we identify several key components that enable us to alleviate expensive inference cost and high parameter requirements while retaining semantic performance at low latencies. 
The main contributions of our work are as follows: We propose HALSIE, a simple and efficient composable architecture with (1) novel hybrid spatio-temporal feature extraction scheme to effectively combine events and frames allowing better information retrieval from a scene (compared to these modalities working independently), (2) and multi-scale cue mixing to enable powerful cross-domain feature integration between the aggregated temporal features and current spatial feature. Our method is lightweight, inference-efficient and still offers state-of-the-art performance for semantic segmentation.
(3) We evaluate HALSIE on real-world DDD-17~\cite{binas2017ddd17}, MVSEC~\cite{zhu2018multivehicle}, and DSEC-Semantic datasets~\cite{sun2022ess} and demonstrate up to $9\%$ improvement over the best performance reported so far with significant energy savings.
In addition, we also provide insights into the various components of our method that contribute to these results.

\section{Related Works}
\label{sec:related}
With event cameras showcasing great potential for semantic segmentation, there have been several efforts in recent years
exploring this emerging research direction. 

Recent works explore using \textbf{stateless ANNs} with
dense event representations, discarding temporal correlation across the event window by representing them as channels. Initial work to adapt events for semantic segmentation in~\cite{alonso2019ev} used an Xception-type network~\cite{chollet2017xception} to achieve robust performance in corner case scenes suffering from over-exposure. The authors published the first event-based segmentation dataset with semantic labels~\cite{binas2017ddd17} generated 
on synchronised grayscale frames from DAVIS346B~\cite{brandli2014240, lichtsteiner}. 
Researchers in~\cite{gehrig2020video} showed improvements over~\cite{alonso2019ev} by training on an augmented dataset comprising real and synthetic events converted from videos. However, they require video datasets, very few of which exist for the task. Improving upon their approach, authors in~\cite{wang2021evdistill} attempt to exploit knowledge learned from high-quality labeled image datasets such as Cityscapes~\cite{cordts2016cityscapes} for unpaired event data,
and report better performance.
However, their knowledge distillation process leads to much higher compute costs. In contrast,~\cite{wang2021dual} relies on event-to-image transfer but fails to consider any network blocks to address the inherent temporal correlation in events. Instead, ~\cite{messikommer2022bridging} reports a method for `image-to-event transfer' that splits the embedding space into motion-specific features shared by events and images using adversarial learning. However, their method depends on hallucination of motion from images to generate fake events 
and is prone to mode collapse~\cite{thanh2020catastrophic}. 
\begin{figure}[t]
  \centering
   \includegraphics[width=1.00\linewidth, height=0.4\linewidth]{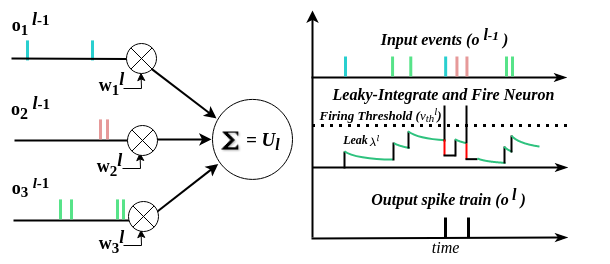}
   \vspace{-6mm}
   \caption{\textbf{Dynamics of a Leaky-Integrate and Fire (LIF) neuron.} The firing threshold $v_{th}^l$ and the leak factor $\lambda^l$ are dynamically updated during training to attain best possible performance.}
   \vspace{-6mm}
   \label{fig:lif}
\end{figure}

A second research direction uses \textbf{temporal recurrence in dense neural networks} to achieve better semantic performance with events. Efforts have been made in event-to-image reconstruction methods, with authors in ~\cite{rebecq2019high} using a recurrent network to convert events to images and processing them with standard ANNs. However, not only do they end up discarding temporal ordering in the inputs in their effort to synthesize motion-invariant images but also incur significant parametric overheads and inference-latency during image generation from events. Follow-up work by~\cite{sun2022ess} attempts to transfer the task from labeled image datasets to unpaired events by aligning similar reconstructed event embeddings generated using the recurrent encoder from~\cite{rebecq2019high} with standard image embeddings. While mostly suffering from the same bottlenecks as~\cite{rebecq2019high}, another significant downside of using dense recurrence comes with synchronous networks like RNNs that struggle with efficient processing when sampling rates of the input are highly variable and asynchronous~\cite{Baltrušaitismultimodal, neil2016phased}.
Typical sub-optimal solutions involve padding, copying, or sampling rate conversions of event frames, which comes at the cost of frequency reductions and temporal misalignments of input data~\cite{oviatt2018handbook}.  

A recent third line of work employed \textbf{SNNs} to propagate information sparsely within
the network. Authors in~\cite{kim2022beyond} were the first to explore a fully-spiking approach, but reported sub-optimal performance compared to state-of-the-art dense networks due to the very sparse
event-driven \emph{binary} compute across their deep ANN-inspired architectures~\cite{chen2017deeplab, chen2014semantic}. 
We build on this line of work to instead propose a hybrid method unifying the advantages of SNNs and ANNs in a compact way. Our method differs from existing work in a few key points: (1) we maximize accumulated \emph{analog} temporal context from event streams while still leveraging sparse implicit recurrence. To achieve this, we build a shallow SNN-based Temporal Feature Extractor (TFE) module that does not rely on any of the earlier additional parametric and learning overheads.
(2) Enhancing the complementary dense spatial embeddings from the ANN-based Spatial
Feature Extraction (SFE) module with aggregated temporal features from the SNN-based TFE module.
(3) We combine the rich spatio-temporal embeddings using a Multi-scale Mixer (MMix) to build a fast, lightweight and highly performant end-to-end learning framework deployable on hybrid ANN-SNN neuromorphic chips such as~\cite{pei2019towards}.

\section{Method}
\begin{figure*}[!ht]
  \centering
    \includegraphics[width=1.0\linewidth, height= 0.41\linewidth]{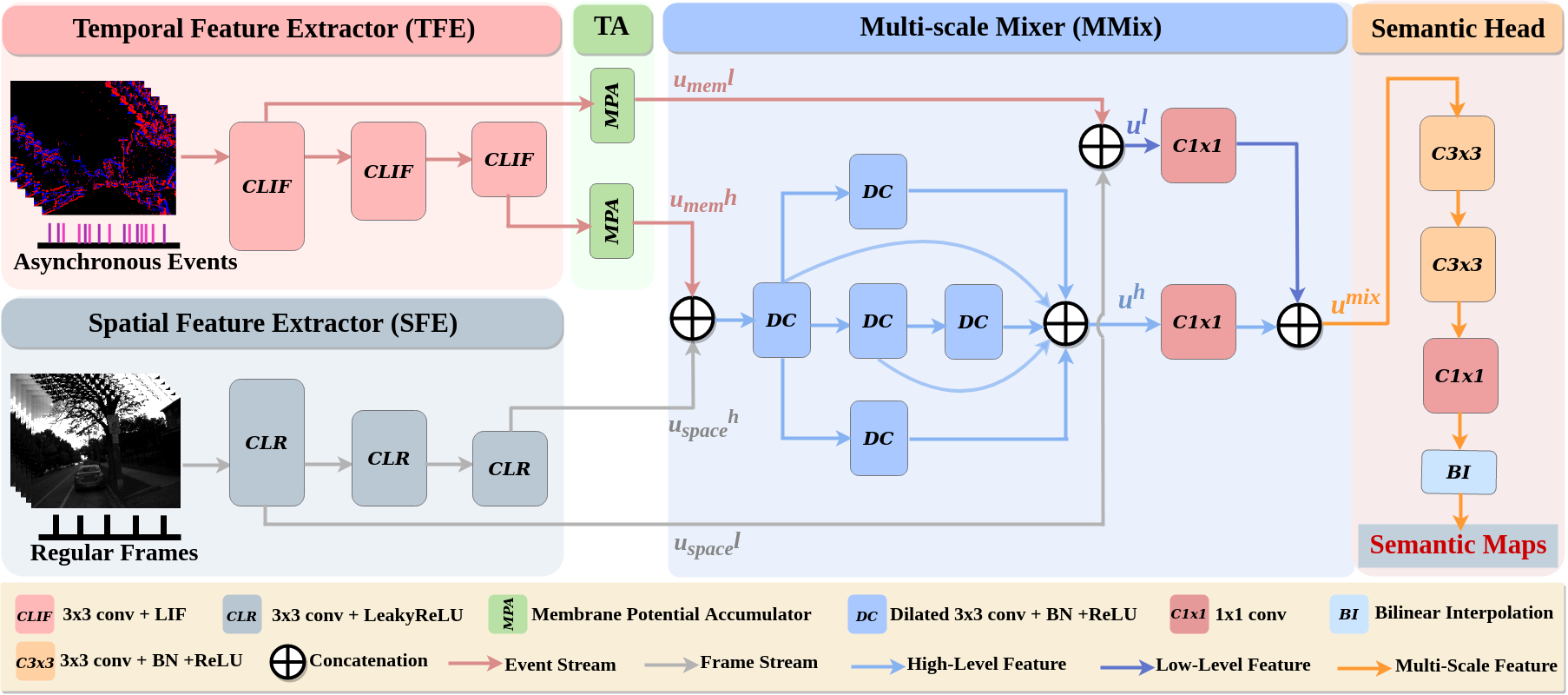}
    \vspace{-6mm}
    \caption{\textbf{Overview} of the proposed HALSIE framework. Given a set of inputs, the TFE and SFE blocks extract rich temporal and spatial embeddings. Temporal Accumulor (TA) and Multi-Scale Mixer (MMix) modules combine analog $u_{mem}$ and $u_{space}$ features by specially designed feature-mixing scheme. Finally, the MMix block interfaces with the segmentation head to generate dense semantic maps.}
    \vspace{-6mm}
    \label{fig:arch}
  \hfill
\end{figure*}

Our segmentation approach is designed to process a stream of events sequentially as they arrive. In every timestep, our network takes
a new event bin as input and relies on the accumulated neuronal state from previous inputs to produce temporal feature maps. After mixing the current spatial features and aggregated temporal features, the fused embeddings are used as input to the decoder. Fig.~\ref{fig:arch} shows an
overview of the HALSIE architecture.
\subsection{Input Processing}
We characterize event data in the Address Event Representation (AER) format as a tuple
$e_i = (x_i, y_i, t_i, p_i)$ that occurs at pixel $(x_i, y_i)$ at time $t_i$, and with polarity $p_i \in {0, 1}$. In this work, we employ a simple yet effective pre-processing method to map events into a grid-like presentation
. Our preprocessing step starts with discretizing an aggregated event volume as follows: For a set of $N$ input events $\{(x_i, y_i, t_i, p_i)\}_{i\in[1, N]}$ between two consecutive grayscale images and a set of $B$ event bins to be created within this event volume, we generate discretized event bins using bilinear sampling kernels $k_b(a)$ ~\cite{jaderberg2015}:
\small
\begin{equation}
    t_i^* = (B-1)(t_i-t_1)/(t_N-t_1) \\ 
\end{equation}
\begin{equation}
    V(x,y,t) = \sum_i{p_i k_b (x-x_i) k_b(y-y_i) k_b(t-t_i^*)} \\
\end{equation}
\begin{equation}
    k_b = max[0, 1-|a|]\\ 
\end{equation}
\normalsize
In words, we generate $B$ temporal bin tensors, each having ON/OFF polarity channels containing the number of positive or negative events within each bin, passed sequentially as timesteps through the TFE module to preserve the inter-bin temporal correlation. The intuition behind a multi-channel representation is to allow the network to learn pixel ownership for moving objects (pixels on the same object will move in the same direction, and generate spatially close iso-polarity events) while capturing short-term temporal correlation over timesteps. We use GT-labels on the latter grayscale image in the event window as done in prior art~\cite{sun2022ess, alonso2019ev, wang2021dual, wang2021evdistill}. Other more complex representations exist~\cite{wang2019event, baldwin2022time, barchid2022bina}, but their thorough evaluation is not our focus.

\subsection{Mixing Spatial and Temporal Features}
The HALSIE architecture features a deep hybrid encoder-decoder network for end-to-end learning. To efficiently extract rich spatiotemporal (ST) features from the complementary sensors, we design a dual-path encoder comprising an ANN-based \emph{Spatial Feature Extractor (SFE)} for frames and an SNN-based \emph{Temporal Feature Extractor (TFE)} for events. We enable incorporating a higher effective receptive field by using a $3\times 3$ convolution with overlapping kernels in both encoders that at the same
time spatially decimates the feature map from the previous encoder step (number of channels is scaled up by a factor of 2).

\vspace{2mm}
\hspace{-4mm}\textbf{Temporal Feature Extraction}  \hspace{2mm}  We opt for spike-based temporal feature aggregation with Leaky-Integrate-and-Fire (LIF)~\cite{abbott1999lapicque, dayan2001theoretical} neurons at each layer in the TFE. LIFs are amongst most widely used bio-inspired spiking neuron models because of their inherent ability to `remember' and `recall' past information, skipping computation on neuromorphic hardware if they haven’t received any input event (event-driven computation). 
We characterize the internal dynamics of our LIF neuron model as follows:
\small
\begin{equation}
    u_{mem}^{l}[t] = w^{l}o^{l-1}[t] +\lambda^{l}u_{mem}^{l}[t-1]-v_{th}^{l}o^{l}[t-1]\\
    \label{eq1}
\end{equation}
\begin{equation}
o[t] = \mathcal{H}(u_{mem}[t] - v_{th})
\end{equation}
\normalsize

where $\mathcal{H}$ represents the Heaviside step function~\cite{weisstein2002heaviside}. At timestep t, weighted output spikes from the previous neuron $l-1$ are accumulated in the membrane potential $u_{mem}^l[t]$ of the neuron $l$ creating a `short-term memory'. At the same time, $u_{mem}^l[t]$  of the neuron $l$ decays by a leak factor $\lambda^l$  to represent `forgetting'. Once the accumulated membrane potential exceeds the firing threshold $v^{l}_{th}$, the neuron generates a binary spike output ($o^l$). $u_{mem}^l[t]$ is reset using the `soft reset' strategy~\cite{ledinauskas2020training, han2020rmp} after all the $B$ temporal bins are processed. We regard this sparse potential accumulation, decay, and resting process as an efficient temporal memory, motivating us to investigate the SNN layers for temporal feature extraction. We examine this aspect in~\cref{sfevstfe}. The LIF neural dynamics is shown in Fig.~\ref{fig:lif}.

Binary spike trains are used to transmit temporal features in traditional SNNs. Differently from prior work in Spiking-Deeplab~\cite{kim2022beyond}, we use the analog membrane potential to transmit neural activations from the TFE instead of the generated output spikes. This enables the aggregated temporal embeddings to have higher expressibility than simple binary spikes while still demanding drastically lower compute compared to traditional RNNs. In other words, applying accumulated analog $u_{mem}$ maps from the Temporal Accumulator (TA) module instead of binary spike maps allows our model to avoid the performance degradation experienced in~\cite{kim2022beyond} while still retaining all the benefits of spike-based processing.
Dynamic threshold schemes observed in different biological systems play an essential role in their robustness to various external conditions~\cite{fontaine2014spike, pozo2010unraveling}. Specifically inspired by~\cite{rathi2021diet}, we also learn the $v_{th}$ and $\lambda$ for each layer dynamically to find the optimal hyperparameters unlike methods in~\cite{kim2022beyond}. We find that allowing the network to learn on the go how `important' a new input is, or how much of the past experiences influence its current state by adjusting its threshold and leak enables our dynamical TFE to attain substantial boosts in dense classification tasks. 

\begin{table*}
\caption{\textbf{Comparison on test set of DDD-17, measured by accuracy and mIoU.} Best results in \textbf{bold} and second best \underline{underlined}. Parameter efficiency and inference energy cost is computed on standard 45nm CMOS process~\cite{mac} (See suppl. material for details).}
\vspace{-6mm}
\begin{center}
\resizebox{1.0\textwidth}{!}{
\begin{tabular}{lccccccc}
\hline \hline
 Methods & Accuracy [\%] & mIoU [\%] & Network & $\text{Params} (\times 10^6)$ & $\# \text{FLOPs}_\text{ANN}(\times 10^9)$ & $\# \text{FLOPs}_\text{SNN}(\times 10^9)$ & $\text{E}_\text{Total} (mJ)$ \\ \hline \vspace{-3mm} \\
EV-SegNet~\cite{alonso2019ev} & 89.76 & 54.81 & ANN & 29.09 & 73.62& - & 338.65\\\vspace{-3mm} \\
EvDistill~\cite{wang2021evdistill} & - & 58.02 & ANN & 59.34 & 12.45 & - & 57.27\\ \vspace{-3mm} \\
DTL~\cite{wang2021dual}  & - & 58.80 & ANN & 60.48 & 16.74 & - & 77.01\\\vspace{-3mm} \\
Spiking-Deeplab~\cite{kim2022beyond} & - & 33.70 & SNN & \underline{4.14} & - & \underline{54.34} & 48.91\\\vspace{-3mm} \\
E2ViD~\cite{rebecq2019high} & 83.24 & 44.77 & ANN & 10.71 & 16.65 & - & 76.59\\\vspace{-3mm} \\
EV-Transfer~\cite{messikommer2022bridging} & 47.37 & 14.91 & ANN &  7.37 & \underline{7.88} & - & \underline{36.25}\\\vspace{-3mm} \\
ViD2E~\cite{gehrig2020video} & 90.19 & 56.01 & ANN & 29.09 & 73.62 & - & 338.65\\\vspace{-3mm} \\
ESS~\cite{sun2022ess} (E) & \underline{91.08} & \textbf{61.37} & ANN & 12.91 & 14.22 & - & 65.41\\\vspace{-3mm} \\
ESS~\cite{sun2022ess} (E+F) & 90.37 & 60.43 & ANN & 12.91 & 14.22 & - & 65.41\\\vspace{-3mm} \\
\hline
Ours (HALSIE) & \textbf{92.50} & \underline{60.66} & Hybrid & \textbf{1.82} & \textbf{3.84} & \textbf{0.267} & \textbf{17.89}\\ \hline \hline \vspace{-5mm} 
\end{tabular}}
\end{center}
\vspace{-5mm}
\label{ddd17table}
\end{table*}
\vspace{3mm}
\hspace{-4mm}\textbf{Spatial Feature Extraction}  \hspace{2mm} The ANN-based SFE branch adopts channel-wise dependencies to extract rich texture
cues, which we call spatial potential maps $u_{space}$, from  synchronized grayscale images of the DAVIS sensor temporally closest to the event bins. If multiple such images are available over a temporal window, they can be fed as separate channels at the input. Each SFE block comprises a conv. layer with overlapping kernels, batch-norm
(BN)~\cite{ioffe2015batch} and a LeakyReLU activation~\cite{xu2015empirical}. We further study the SFE branch as part of our architecture variation
in the ablation studies in~\cref{sfevstfe}.

\subsection{Multi-scale Mixer}

In the subsequent step, resulting temporal and spatial embeddings, $u_{mem}$ and $u_{space}$, are mixed using a multi-scale integrator. Spatial feature maps from the SFE at the highest-level and lowest-level scales ($u_{space}^h$ and $u_{space}^l$) are combined with the corresponding analog accumulated membrane potential maps ($u_{mem}^h$ and $u_{mem}^l$) from the Temporal Accumulator (TA) module to obtain high- and low-level mixed potential maps. High-level mixed maps (i.e, $u^h$) from the last encoder layer pass through a \emph{Multi-scale Mixer (MMix)} block with each branch of the cell employing
decoupled rate $3\times 3$ dilated convolutions~\cite{chen2018encoder}. By enabling different sampling rates $r_h \times r_w$ for each dilated conv. cell, we capture object scales with different aspect ratios, and create a more diverse feature space with each branch of the block building local multi-scale contextual information through parallel or cascaded representations. More details regarding the decoupled sampling rates for the model can be found in the supplementary. 
As a next step, the low-level $u^l$ maps and multi-scale high-level $u^h$ maps are concatenated after channel-mixing ($1\times 1$ or pointwise convolutions) corresponding to global, dilated mixed features $u^{mix}$. 

\subsection{Semantic Head}
For the semantic head, we adopt a lightweight task decoder consisting of 2 [(3$\times$3 conv) → (BN) → (ReLU)] blocks followed by a (1$\times$1 conv) and upsampling layer to predict the segmentation mask. We examine the $u^{mix}$ feature maps as a toolkit to visualise and interpret why such a simple decoder design works well for our method and discuss results in~\cref{sfevstfe}.
\section{Experiments and Evaluation}
\label{sec:exp}
\subsection{Setup}
Our models are trained $100$ epochs with the ADAM optimizer~\cite{kingma2015adam} using a MultiStepLR learning rate schedule to scale the learning rate by
0.7 every 10 epochs. We use a weighted pixel-wise cross entropy loss to examine each pixel individually. Unlike standard backpropagation in ANNs, gradient computation in SNNs is not straightforward since LIF neurons have a spiking mechanism that generates non-differentiable threshold functions. We enable learning with surrogate gradients to approximate the gradient of the Heaviside step function during backpropagation~\cite{neftci2019surrogate, lee2016training} in our TFE branch and use the inverse tangent surrogate gradient function with width $\gamma = 100$ (to allow sufficient gradient flow) since it is computationally inexpensive. To construct event representations, we discretize the event window between consecutive frames into $B = 10$ temporal bins bins and pass them along with the synchronized grayscale frames to the TFE and SFE branches respectively. To estimate energy costs for a single inference, we use the number of floating point operations (FLOPs) performed by the network per inference cycle. For details on computing approximate inference energy, refer to the supplementary material.
\subsection{Evaluation on DDD-17 Dataset}
\begin{figure*}[!ht]
  \centering
    \includegraphics[width=1.0\linewidth, height=0.33\linewidth]{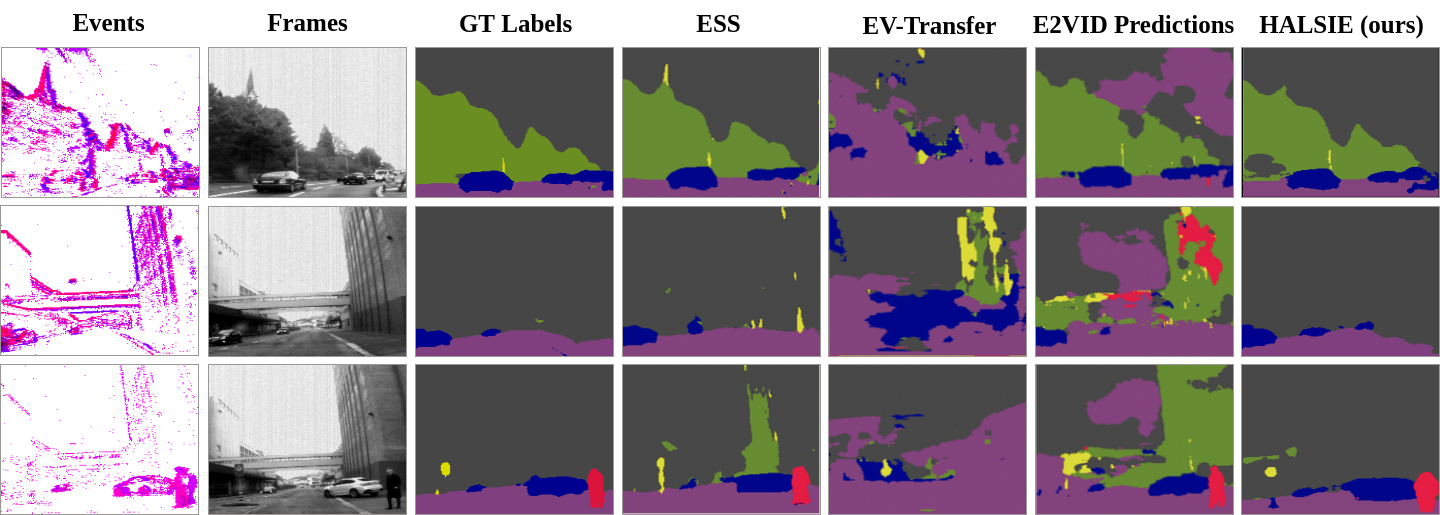}
    \vspace{-6mm}
    \caption{\textbf{Samples on DDD-17.} Compared to prior art, our highly lightweight hybrid framework generates more reliable predictions with upto $72.7\%$ lower inference energy (gray: background; green: vegetation;
blue: vehicle; violet: street; yellow: object; red: person)}
    \vspace{-6mm}
    \label{fig:ddd17_comp}
  \hfill
\end{figure*}
\hspace{-5pt}\textbf{ Dataset and Training Details:} We use the publicly available driving scene dataset DDD-17~\cite{binas2017ddd17}, containing $40$ different driving sequences of synchronized grayscale images and event streams. 
Due to the low resolution of the DAVIS camera, several classes are fused to create labels for six merged classes. From the provided sequences, this work uses a training set of $15,584$ frames and a test set of $3,584$ frames. Maintaining parity with prior art~\cite{alonso2019ev, gehrig2020video, wang2021evdistill}, we use constant integration time event bins with $T =50ms$.  
Our data augmentation includes random flips and rotations on inputs and cropping them to $192 \times 192$ size images. 
We train batch sizes of $32$ on an initial learning rate of $8e-4$. and report accuracy and mean intersection over union (mIoU) on our semantic maps to evaluate performance. 

\hspace{-5pt}\textbf{Results:}
Quantitative results are reported in Table~\ref{ddd17table} and visualized in Fig.~\ref{fig:ddd17_comp}. We compare our approach with existing works such as~\cite{alonso2019ev, wang2021evdistill, messikommer2022bridging, wang2021dual, gehrig2020video} that do not leverage temporal correlation between events, and find that our hybrid framework leverages the complementary events and frames with efficient spatio-temporal learning to consistently outperform them and achieve new state-of-the-art performance of $60.66\%$ mIoU and $92.50\%$ accuracy. ESS~\cite{sun2022ess} which uses a recurrent event encoder to hallucinate motion-invariant event embeddings in the event only (E) and event+frame (E+F) settings claims comparable results, albeit at the cost of $85.9\%$ lower parameter efficiency. This can be mainly attributed to the inherent self-recurrence in SNNs which are more suitable to denoise and extract sparse cues compared to traditional RNNs which are not designed for sparse, asynchronous or irregular data (also pointed out in~\cite{deng2020rethinking}). We will verify this later in~\cref{sfevstfe} and the supplementary. Our lightweight model is the smallest in our comparison with up to a staggering $73\%$ lower inference cost than existing approaches. Still, our efficient multi-scale cross-domain feature mixing allows our method to generate the most reliable predictions without compromising on qualitative performance, visualized in Fig.~\ref{fig:plot}.

We also observe that the fully-spiking approach in ~\cite{kim2022beyond} using reconfigured LIF neurons to include atrous convolutions reports very low performance, in line with our intuition regarding the need for multi-modal hybrid networks to leverage complementary information from both sensors. In the first row in Fig.~\ref{fig:ddd17_comp}, our semantic maps were unable to predict the tower peak above the vegetation and classifies it as vegetation itself. However, since the tower peak is not a crucial element in the scene compared to the presence of a nearby traffic pole, an incoming vehicle, or a person, we posit that the error is not critical. Our method makes more reliable predictions in the scenes in the second and third rows compared to its counterparts. 

\subsection{Evaluation on MVSEC Dataset}

\begin{figure*}[t]
  \centering
    \includegraphics[width=1.0\linewidth, height=0.22\linewidth]{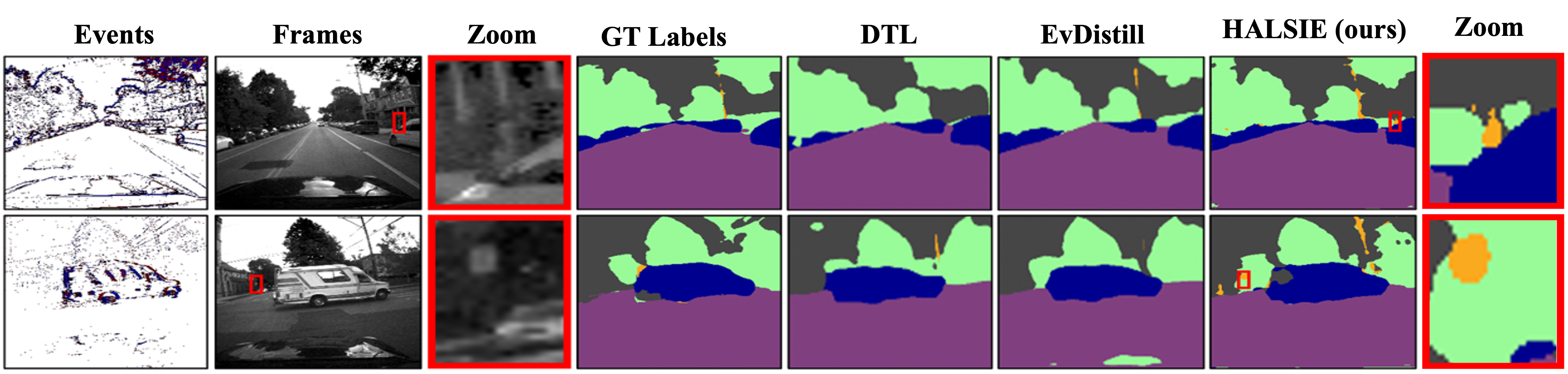}
    \vspace{-6mm}
    \caption{\textbf{Results on MVSEC.} Small objects are sometimes missed in the low-quality semantic labels (zoomed-in patch in the $3^{rd}$ column). This leads to a lower detection score on MVSEC even though our predictions provide more fine-grained detections (see $8^{th}$ column). (gray: background; green: vegetation; blue: vehicle; violet: street; yellow: object)}
    \vspace{-4mm}
    \label{fig:mvsec_comp}
  \hfill
\end{figure*}

\hspace{-5pt}\textbf{Dataset and Training details:}  As events in the DDD-17 dataset are very sparse and noisy, we present experimental results on the MVSEC dataset~\cite{zhu2018multivehicle} comprising of various driving scenes for stereo estimation. Due to the poor quality of frames in the `outdoor day1' sequence, we mainly use the `outdoor day2' sequence and divide data into training and testing sets~\cite{wang2021dual, wang2021evdistill}. We also remove redundant sequences such as vehicles stopping at traffic lights, etc. and train with batch sizes of $32$ and an initial learning rate of $8e-4$. 
\begin{table}[t]
\caption{\textbf{Comparison on MVSEC dataset}. Existing approaches~\cite{wang2021evdistill} and~\cite{wang2021dual} fail to report their accuracy metrics.}
\vspace{-5mm}
\begin{center}
\resizebox{1.0\linewidth}{!}{
\begin{tabular}{lccccc}
\hline \hline
 Methods & Accuracy [\%] & mIoU [\%] & $\text{Params} (\times 10^6)$ & $\text{E}_\text{Total} (mJ)$ \\ \hline \vspace{-3mm} \\
EvDistill~\cite{wang2021evdistill} & - & 55.09  & \underline{59.34} &\underline{101.84}  \\ \vspace{-3mm} \\
DTL~\cite{wang2021dual}  & - & \underline{60.82} & 60.48 & 136.89  \\\vspace{-3mm} \\
\hline
Ours (HALSIE) & \textbf{92.13} & \textbf{66.31} & \textbf{1.82} & \textbf{31.39}\\ \hline \hline \vspace{-5mm} 
\end{tabular}}
\end{center}
\vspace{-3mm}
\label{mvsectable}
\end{table}
\begin{table}
\caption{\textbf{Results on DSEC-Semantic.} Our method shows comparable performance to ESS~\cite{sun2022ess} with a staggering $74\%$ lower inference cost, making it a prime candidate for edge-deployment.}
\vspace{-2mm}
\centering
\resizebox{1.0\linewidth}{!}{
\begin{tabular}{lccccc}
\hline \hline
 Methods & Accuracy [\%] & mIoU [\%] & $\text{Params} (\times 10^6)$ & $\text{E}_\text{Total} (mJ)$ \\ \hline \vspace{-3mm} \\
EV-Transfer~\cite{messikommer2022bridging} & 60.50 & 23.20 & \underline{7.37} & \underline{197.48}\\\vspace{-3mm} \\
E2ViD~\cite{rebecq2019high} & 76.67 & 40.70 & 10.71 & 416.99 \\\vspace{-3mm} \\
EV-SegNet~\cite{alonso2019ev} & 88.61 & 51.76 & 29.09 & 1863.83\\\vspace{-3mm} \\
ESS~\cite{sun2022ess} (E+F) & \textbf{89.37} & \textbf{53.29} & 12.91 & 356.32\\\vspace{-3mm} \\
\hline
Ours (HALSIE) & \underline{89.01}  & \underline{52.43} & \textbf{1.82}& \textbf{94.41}\\ \hline \hline \vspace{-3mm} 
\end{tabular}}
\vspace{-5mm}
\label{dsectable}
\end{table}
\begin{figure}[!ht]
  \centering
    \includegraphics[width=1.0\linewidth, height=0.75\linewidth]{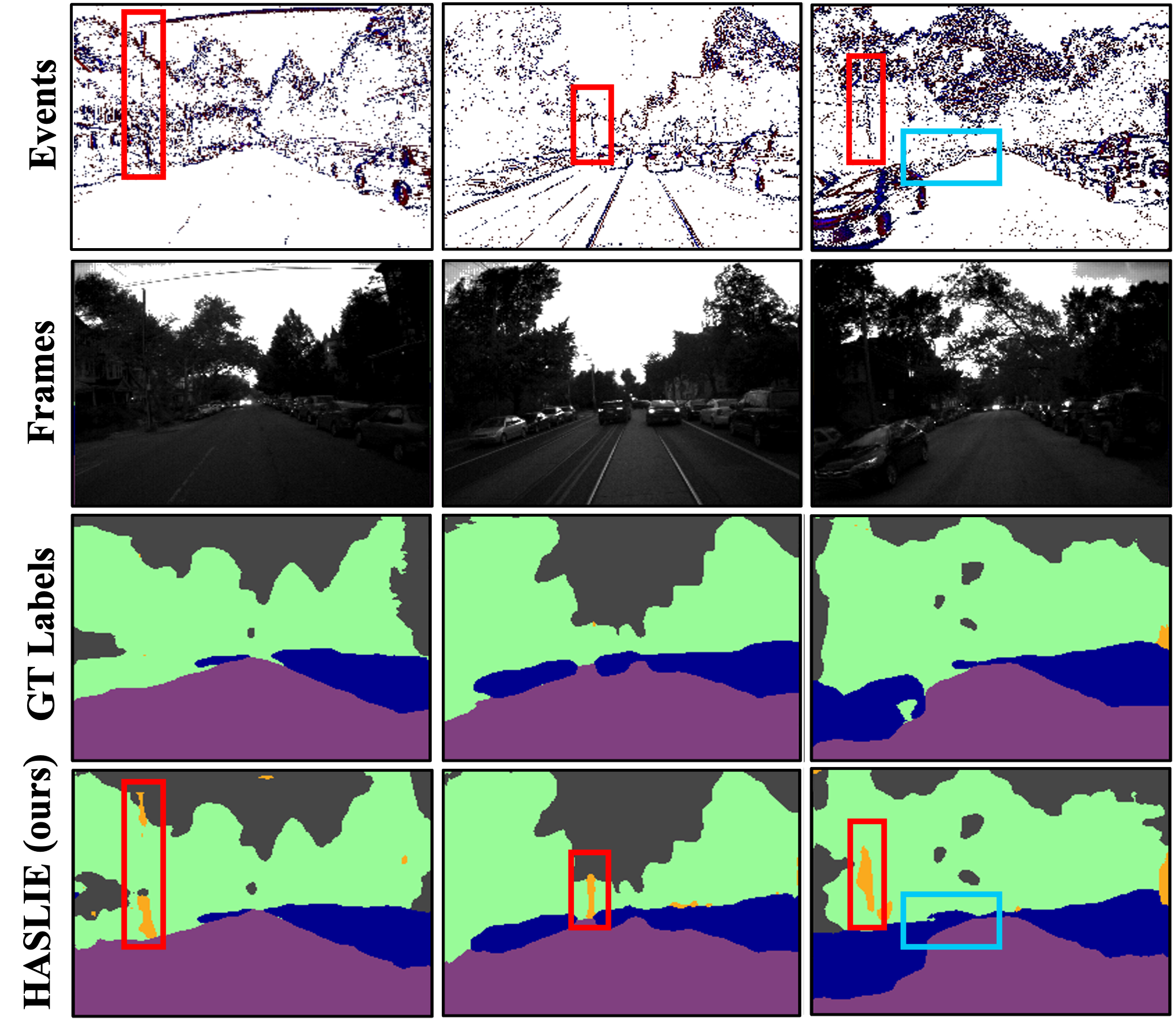}
    \vspace{-7mm}
    \caption{\textbf{Results on challenging scenes.} Qualitative samples of HDR scenes in the MVSEC dataset. Best viewed in color.}
    \vspace{-7mm}
    \label{fig:hdr}
  \hfill
\end{figure}

\hspace{-5pt}\textbf{Results:}
We summarize the results in Table~\ref{mvsectable}. HALSIE outperforms the two existing methods ~\cite{wang2021evdistill} and~\cite{wang2021dual} by around $20.4\%$ and $9.1\%$ respectively, while using $33\times$ fewer parameters and up to a significant $77\%$ lower inference energy.
Effectiveness of our method can also be verified from Fig.~\ref{fig:mvsec_comp} where our segmentation results are able to detect very fine details such as a poles or traffic signs ($8^{th}$ column), where our contemporaries fail. The results validate that our highly lightweight framework is able to efficiently leverage spatiotemporal context from both modalities to predict dense semantic maps with fine predictions. In several examples in Fig.~\ref{fig:mvsec_comp}, we found that our method predicts objects which were not present in the GT labels but were clearly visible in the images, causing misleading reductions in our detection score. We also use HDR scenes to show how our method provides reliable performance when grayscale frames are ill-exposed. HALSIE is able to efficiently extract information from events and shows promising performance in such challenging conditions where fine details are almost invisible in the grayscale frames (highlighted in colored boxes). See ~\cref{fig:hdr} for qualitative results.

\subsection{ Evaluation on DSEC-semantic Dataset}
\hspace{-5pt}\textbf{Dataset and Training Details:} We further evaluate our method on the recently released DSEC-semantic~\cite{sun2022ess} dataset containing $4017$ training and $1395$ testing samples with $11$ semantic classes~\cite{gehrig2021dsec}. The dataset was collected in a variety of urban and rural environments using automotive-grade standard cameras and high-resolution event cameras. Similar to~\cite{gehrig2021dsec}, we generate $B=10$ event bins with a constant event density of $100K$ events/bin to be passed sequentially to the TFE block and associate event bins with labels using the provided semantic timestamps. The SFE module is fed with images from the left frame-camera corresponding to the same semantic timestamp as the events. We train for $100$ epochs with an initial learning rate of $5e-4$. 

\hspace{-5pt}\textbf{Results:} 
The performance of our method evaluated on the test set is reported in Table~\ref{dsectable}. We compare our approach with~\cite{rebecq2019high, messikommer2022bridging} that do not leverage temporal recurrence and find that HALSIE significantly improves segmentation results, surpassing existing methods with around $29\%$ increase in mIoU while using a significant $77.4\%$ lower inference cost. ESS~\cite{sun2022ess} in the events+frames (E+F) setting claims comparable results, while using a much larger network and exorbitant inference costs ($\sim 73.5\%$ higher). Finally, our model is the smallest amongst existing literature by a large margin and still achieves $1.3\%$ higher mIoU than EV-Segnet~\cite{alonso2019ev} while using $16\times$ fewer parameters, making it a top-bidder for energy-efficient edge-applications. Refer to the suppl. for qualitative samples from our method.

\begin{table}[t]
\caption{\textbf{Architectural and input modality variations.} We denote the single encoder setting as `SE’ and dual encoder setting as
`DE’. All models are approximately iso-parameterized.}
\vspace{-5mm}
\begin{center}
\resizebox{1.0\linewidth}{!}{
\begin{tabular}{lccccc}
\hline \hline
Methods & Training Data & Accuracy [\%] & mIoU [\%]&\\ \hline \vspace{-3mm} \\
\emph{(A)} SE: only SFE & frames & 88.15 & 56.38 \\  \vspace{-3mm} \\
\emph{(B)} SE: only SFE & events & 83.42 & 46.01 \\  \vspace{-3mm} \\
\emph{(C)} SE: only TFE & events & 83.59 & 46.09 \\ \vspace{-3mm} \\
\emph{(D)} DE: both SFE & events + frames & 90.16 & 58.28 \\  \vspace{-3mm} \\
\emph{(E)} DE: both TFE & events + frames & 86.01 & 54.43 \\ \vspace{-3mm} \\ 
\emph{(F)} w/o MSFI & events + frames & 90.31 & 58.83 \\  \vspace{-3mm} \\
\emph{(G)} SNN $\rightarrow$ LSTM & events + frames & 90.25 & 58.67 \\  \vspace{-3mm} \\
\hline
\emph{(H)} (ours) HALSIE & events + frames & \bf{92.50} & \bf{60.66} \\
\hline \hline
\end{tabular}
}
\end{center}
\vspace{-7mm}
\label{table1}
\end{table}
\begin{figure}[!ht]
  \centering
    \includegraphics[width=1.0\linewidth, height=0.56\linewidth]{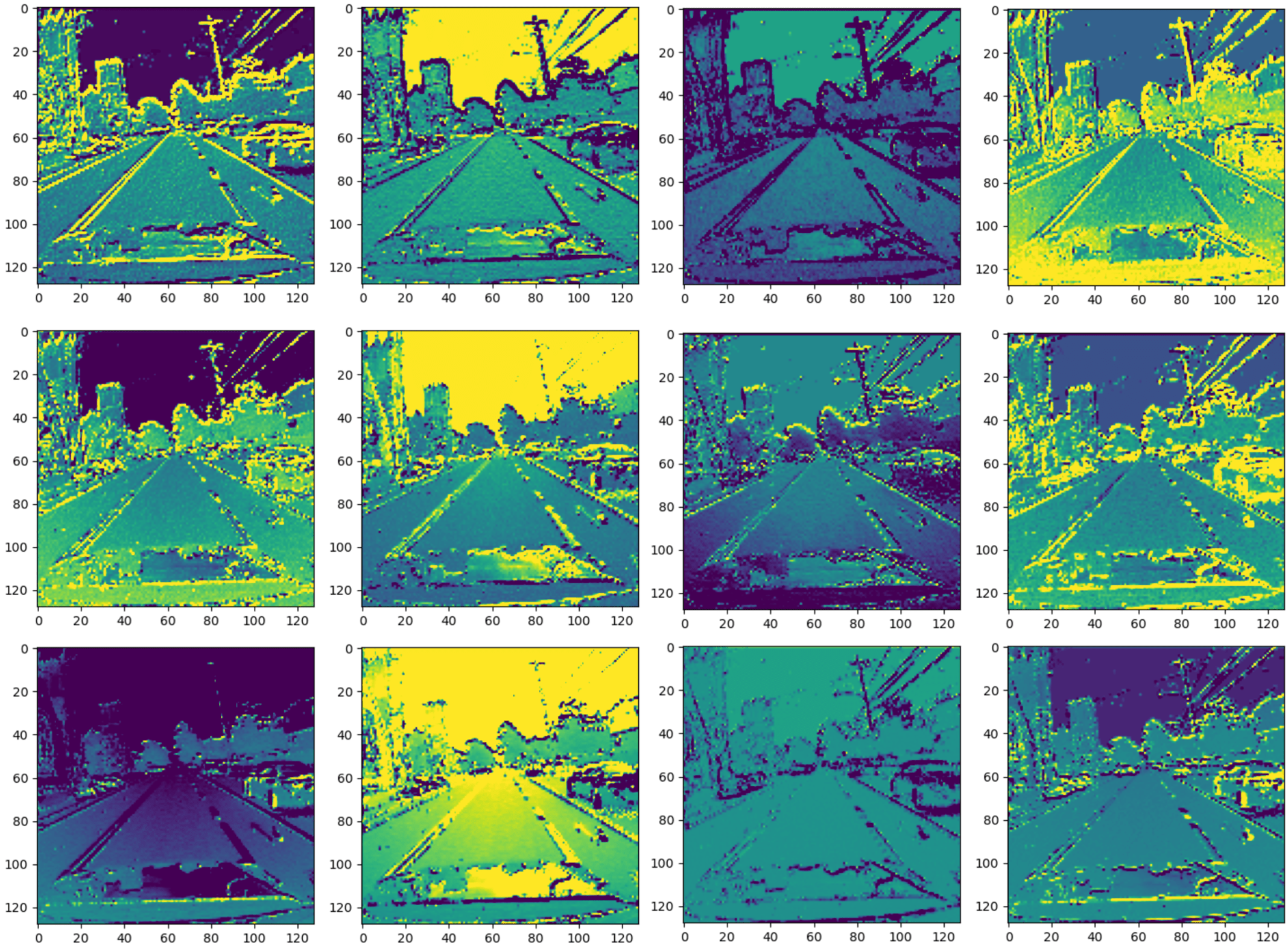}
    \vspace{-6mm}
    \caption{\textbf{Mixed feature maps on MVSEC.} Four channels of the $u^{mix}$ maps generated from processing Top row: Events in TFE and Frames in SFE. Middle row: Both inputs using SFE. Bottom row: Both inputs using TFE. Best viewed with zoom in.}
    \vspace{-7mm}
    \label{fig:feats}
  \hfill
\end{figure}

\subsection{Ablation studies}
\subsubsection{Mixing Spatial and Temporal Cues} \label{sfevstfe}
We conduct the following experiments on DDD-17 to demonstrate effects of spatial and temporal cues on the task: \emph{(A)} and \emph{(B)} Removing the TFE branch but keeping the SFE branch; \emph{(C)} Removing SFE, but keeping TFE; \emph{(D)} Dual encoder setting with SFE branches for both frames and events; \emph{(E)} Same setting but with TFE branches instead; \emph{(F)} Removing the MSFI module; \emph{(G)} Replacing the SNNs in SFE with an LSTM~\cite{shi2015convolutional} with one conv. layer per cell. Comparing the original HALSIE \emph{(H)} to \emph{(A)}, \emph{(B)}, \emph{(C)}, \emph{(D)} or \emph{(E)}, we witness a significant performance degradation, validating that effectively extracting and mixing temporal and spatial cues is essential for boosting performance. Note that setting \emph{(A)} draws comparison of our method with pure frame-based segmentation. Unsurprisingly, when trained with both the modalities, we can detect small objects like traffic poles and people. In several examples, we find that our method segments objects that were not present in the labels, but were clearly visible in the images. For visualization, refer to the supplementary material. 

We use the $u^{mix}$ feature maps as a toolkit to visualise and interpret how our simple decoder benefits from the rich mixed feature representation. We study~\cref{fig:feats} and find that mixed feature maps generated from processing events with TFE and frames with SFE leads to the sharpest and cleanest features (row 1) even with such simple encoders. In contrast, when using SFE for both inputs (row 2), the feature maps appear less discriminative due to the non-sparse processing of events which does not contribute to sharp edge-extraction. Note that these features are however generated at the highest inference cost amongst the three variants due to dense processing in both encoder branches. While applying TFE to both inputs (row 3) offers the most energy-efficient sparse processing paradigm, the feature maps appear relatively noiseless but suffer from loss of information. As such, our simple decoder design does not work as well on the only SFE or only TFE encoder approaches since it is unable to take advantage of the powerful representation induced by complementary SNN-ANN processing. 
\begin{table}[t]
\caption{\textbf{Event representation strategy.} Constant event density (CED) bins lead to best results on the DSEC-Semantic dataset.}
\vspace{-2mm}
\centering
\resizebox{1.0\linewidth}{!}{
\begin{tabular}{l|ccc|ccc}
\hline \hline
 Event  & & CED & & & CIT &\\
Representation & 10k & 100k & 1000K & 10ms & 50ms & 100ms\\\hline 
Accuracy [\%]& 88.78 & 89.01 & 88.24 & 87.76 & 88.02 & 87.23\\
mIoU [\%] & 51.58 & 52.43 & 51.01 & 50.14 & 50.67 & 49.89\\
\hline \hline 
\end{tabular}}
\vspace{-4mm}
\label{dsecablationtable}
\end{table}

We also notice performance degradation in the \emph{(F)} setting. This results reflects the effectiveness of the proposed MSFI block. Compared to \emph{(G)}, we still achieve better performance. This is to some degree surprising because LSTMs can also extract temporal cues. Notably, the spiking mechanism of SNNs acts not only as temporal memory but also as a natural noise filter, which is beneficial to robust predictions. We further examine the denoising aspect of our TFE module in the supplementary material.  

\subsubsection{Event representation and event density}
Thorough evaluation of event representations is not our focus and hence we only study the influence of simple event representations which may not leverage the full potential of event data~\cite{li2022asynchronous} on our method's performance. Efficient low-level encoding of event data is still an open research problem that we have not addressed in this work. Ablation results in Table~\ref{dsecablationtable} and visualisations in the supplementary on the DSEC-Semantic dataset suggest that maintaining a moderately dense bin with constant event density (CED) shows better semantic performance compared to high density bins with trailing artifact events from fast moving objects, or low density bins with minimal contribution to the segmentation performance. We also find that having CED bins consistently helps the network learn the end-task better than with constant integration time (CIT) bins. 


\section{Conclusion}
We introduce HALSIE, a lightweight yet powerful hybrid end-to-end framework for semantic segmentation that is capable of effectively mixing temporal and spatial cues encoded in events and frames. The proposed network relies on several novel modules. We devise an SNN-based temporal feature extractor and an ANN-based spatial feature extractor, which efficiently exploits statistical cues of spatial and temporal information for robust predictions. We also introduce a novel multi-scale mixer for compactly combining embeddings from the two domains. Effectiveness of our design choices is evidenced by the strong performance of our method in detecting finer details on DDD-17, MVSEC and DSEC-Semantic benchmarks while offering sizeable benefits in terms of inference cost and parameter efficiency. The resulting design is deployable for resource-constrained edge applications, and paves the way for low-energy semantic segmentation with event cameras without compromising on performance. Nonetheless, we hope that this work also inspires novel designs in future hybrid systems. 

\section{Acknowledgement}
This work was supported in part by the Center for Brain- inspired Computing (C-BRIC), a DARPA sponsored JUMP
center, the Semiconductor Research Corporation (SRC), the
National Science Foundation, the DoD Vannevar Bush Fellowship, and IARPA MicroE4AI.

\newpage
\section{Supplementary Material}
\label{sec:suppl}
The supplementary material is organized as follows: Section~\ref{sec:suppl-1} investigates the denoising characteristic of SNNs; Section~\ref{sec:suppl-2} discusses edge-feature extraction using ANNs vs SNNs; Section~\ref{sec:suppl-3} highlights qualitative evaluations on the newly released DSEC-Semantic dataset; and Section~\ref{sec:suppl-4} provides details regarding computing approximate inference cost for different methods based on the number of floating point operations they perform per cycle. Section~\ref{sec:suppl-5} elaborates on the decoupled sampling rates used in our architecture. Section~\ref{sec:suppl-6} finally provides some additional visualisations to inspect phenomenon we discuss in the main paper.

\subsection{Denoising with SNNs}
\label{sec:suppl-1}
In order to segment objects efficiently, we aim to isolate relevant events corresponding to objects of interest in the scene, filtering out any spurious event inputs generated due to background clutter and sensor noise. Spiking neurons such as the Leaky-Integrate and Fire are excellent candidates as they are capable of maintaining an internalized state called membrane potential $u_{mem}$, which decays over time at a rate controlled by the leak factor. The leak factor denotes how much of the membrane potential is retained
for the next time step, i.e., the higher the leak factor, the
slower the rate of decay. If the accumulated membrane
potential of the neuron exceeds the threshold at any point,
$(u_{mem} >v_{th})$, the neuron emits an output spike and resets
its membrane potential.

\begin{figure}[t]
  \centering
    \includegraphics[width=1.0\linewidth]{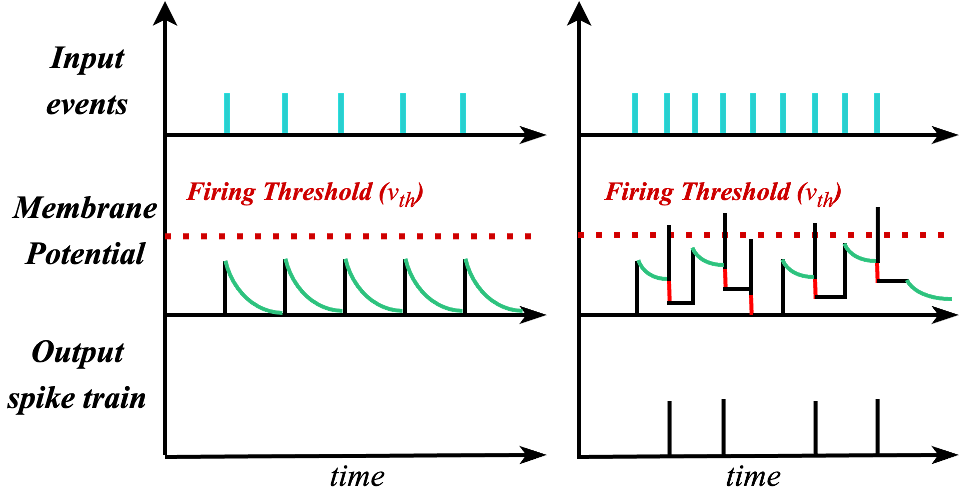}
    \vspace{-7mm}
    \caption{\textbf{Temporal sensitivity of a LIF neuron.} Left graph shows the neuron response to inputs occurring at a lower rate; right graph shows the neuron response to inputs occurring at a higher rate. Once the inputs occur at a rate higher than a certain frequency, the neuron generates spikes and resets to a resting value.}
    \vspace{-7mm}
    \label{fig:denoiselif}
  \hfill
\end{figure}

Spurious events due to sensor noise are usually generated at much lower rates than events triggered by moving vehicles, humans or other relevant objects of interest in a scene. 
As such, if the time between input events
is large, the membrane potential decays its value before it can reach the threshold. However, if these input events occur more frequently, they are able to overcome the decay and increase the membrane potential towards the threshold. Thus, the
neuron generates output spikes if the input events occur at a frequency higher than a certain value. We visualise this phenomenon in~\cref{fig:denoiselif}. Leveraging this sparse spiking property of LIF neurons, our SNN-based TFE module enjoys the benefits of implicit denoising of input events with no extra parametric or learning overheads in contrast to dense ANNs or RNNs which do not directly lend themselves to filtering out sensor noise.
\begin{figure*}[t]
  \centering
    \includegraphics[width=1.0\linewidth, height=0.55\linewidth]{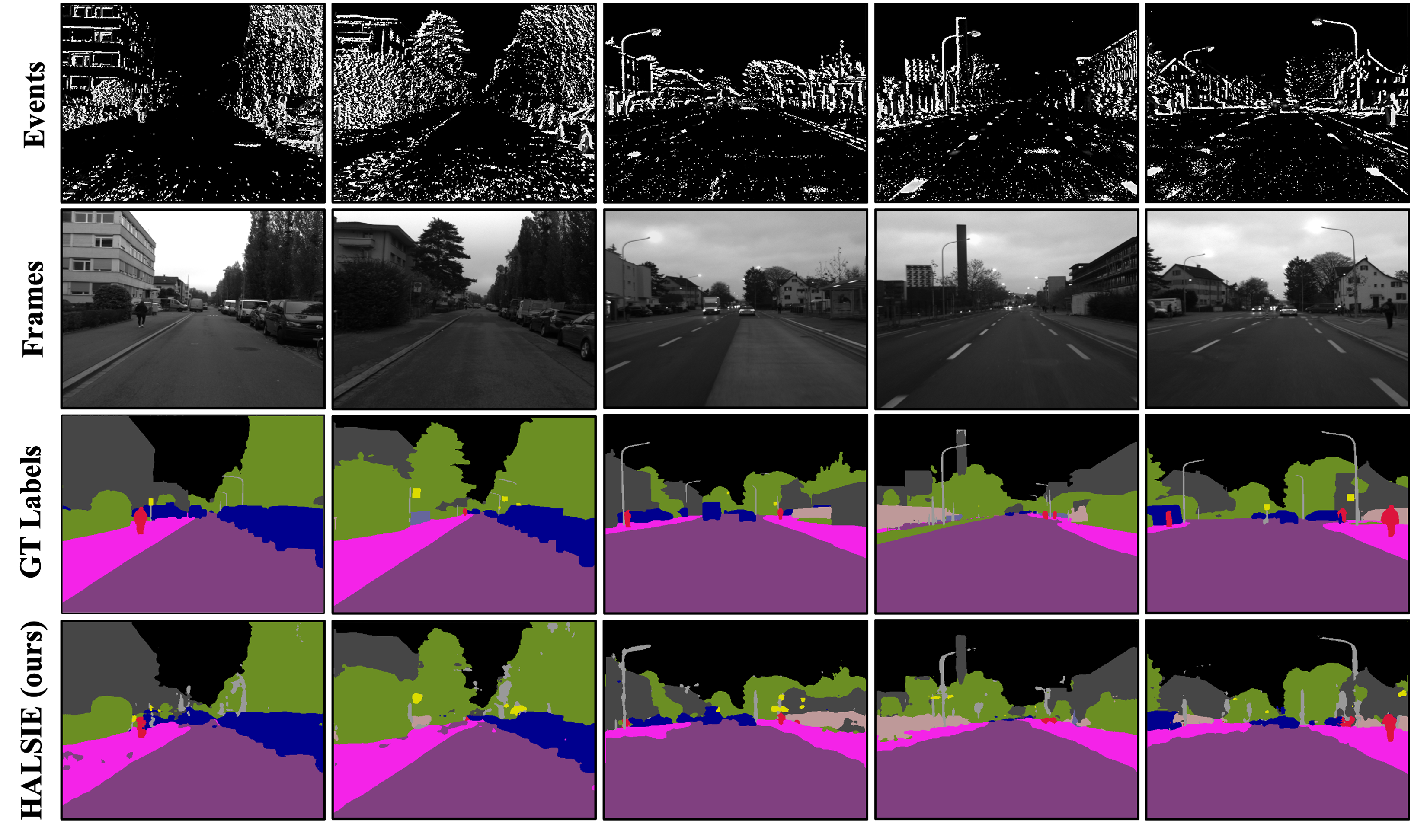}
    \vspace{-7mm}
    \caption{\textbf{Semantic segmentation results on the DSEC-Semantic dataset.} Best viewed in color.}
    \vspace{-5mm}
    \label{fig:2}
  \hfill
\end{figure*}

\subsection{Sharp feature map extraction with ANNs vs SNNs}
\label{sec:suppl-2}
\begin{figure}[t]
  \centering
    \includegraphics[width=1.0\linewidth, height=0.6\linewidth]{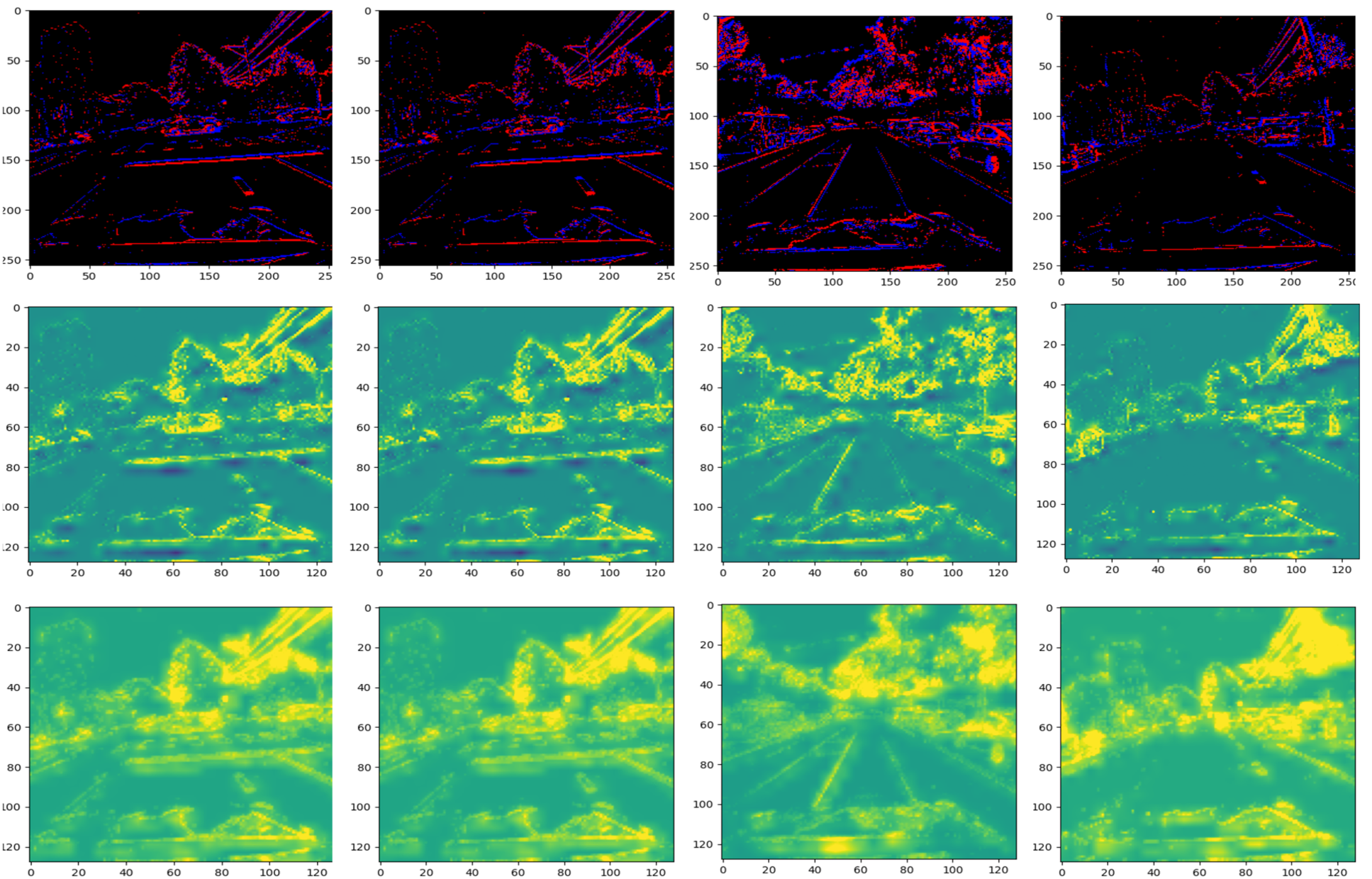}
    \vspace{-5mm}
    \caption{\textbf{Feature maps} generated by processing Top row: events inputs with Middle row: SNN-based TFE and Bottom row: ANN-based SFE.}
    \vspace{-5mm}
    \label{fig:denoise}
  \hfill
\end{figure}

We know that in stable lighting conditions, events are triggered by moving edges (e.g., object contour and texture boundaries), making an event-based camera a natural edge extractor. We try to use this captured edge information and investigate how sparse processing of events in our SNN-based TFE is better suited to not only high energy savings, but also implicitly supports sharp edge extraction. We investigate feature maps from processing events using dense methods like ANNs instead of sparse methods like SNNs. Visualising results in~\cref{fig:denoise}, we find that using SNNs for extracting cues from the input results in edge-discriminative feature maps (row 2) caused by high spike generation at the object-boundary pixels where events are triggered at higher rates. In contrast, using ANN-like dense processing networks (row 3) do not offer similar benefits as is seen in~\cref{fig:denoise} with the feature maps looking fuzzier.
\subsection{Evaluation on DSEC-Semantic Dataset}
\label{sec:suppl-3}
We provide qualitative samples of HALSIE on the DSEC-Semantic dataset~\cite{sun2022ess}. Fig.~\ref{fig:2} shows our model can successfully detect objects in various scenes on the test set comprising `zurich$\_$city$\_$13$\_$a’,
`zurich$\_$city$\_$14$\_$c’ and `zurich$\_$city$\_$15$\_$a’ sequences. Interestingly, even with a complex dataset such as DSEC-Semantic with 11 semantic classes: background, building, fence, person, pole, road, sidewalk, vegetation, car, wall, and traffic sign, our lightweight and inference-efficient method does not fall short and is still able to segment large objects with almost accurate boundaries, and detect smaller objects with fine-grained details in a variety of driving scenes.

\begin{figure*}[!ht]
  \centering
    \includegraphics[width=1.0\linewidth]{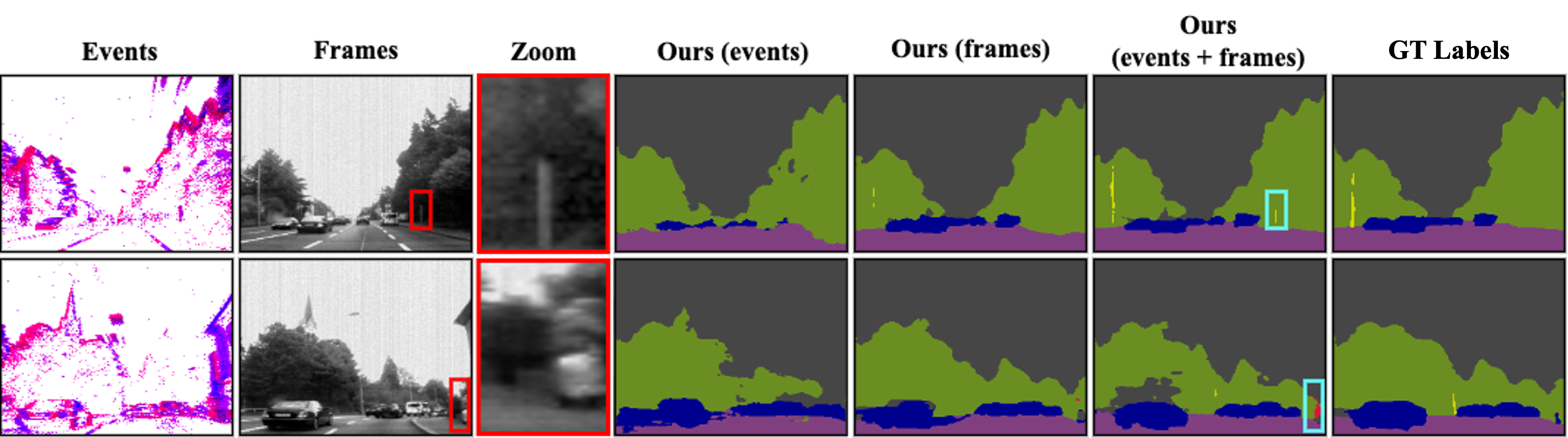}
    \vspace{-5mm}
    \caption{\textbf{Segmentation results for multi-modal vs. uni-modal settings.} Finer objects missing from the GT-labels (zoomed-in patch in the red box) are detected by our hybrid multi-modal method trained using events + frames.}
    \vspace{-5mm}
    \label{fig:fused}
  \hfill
\end{figure*}
\begin{figure}
  \centering
    \includegraphics[width=1.0\linewidth, height=0.78\linewidth]{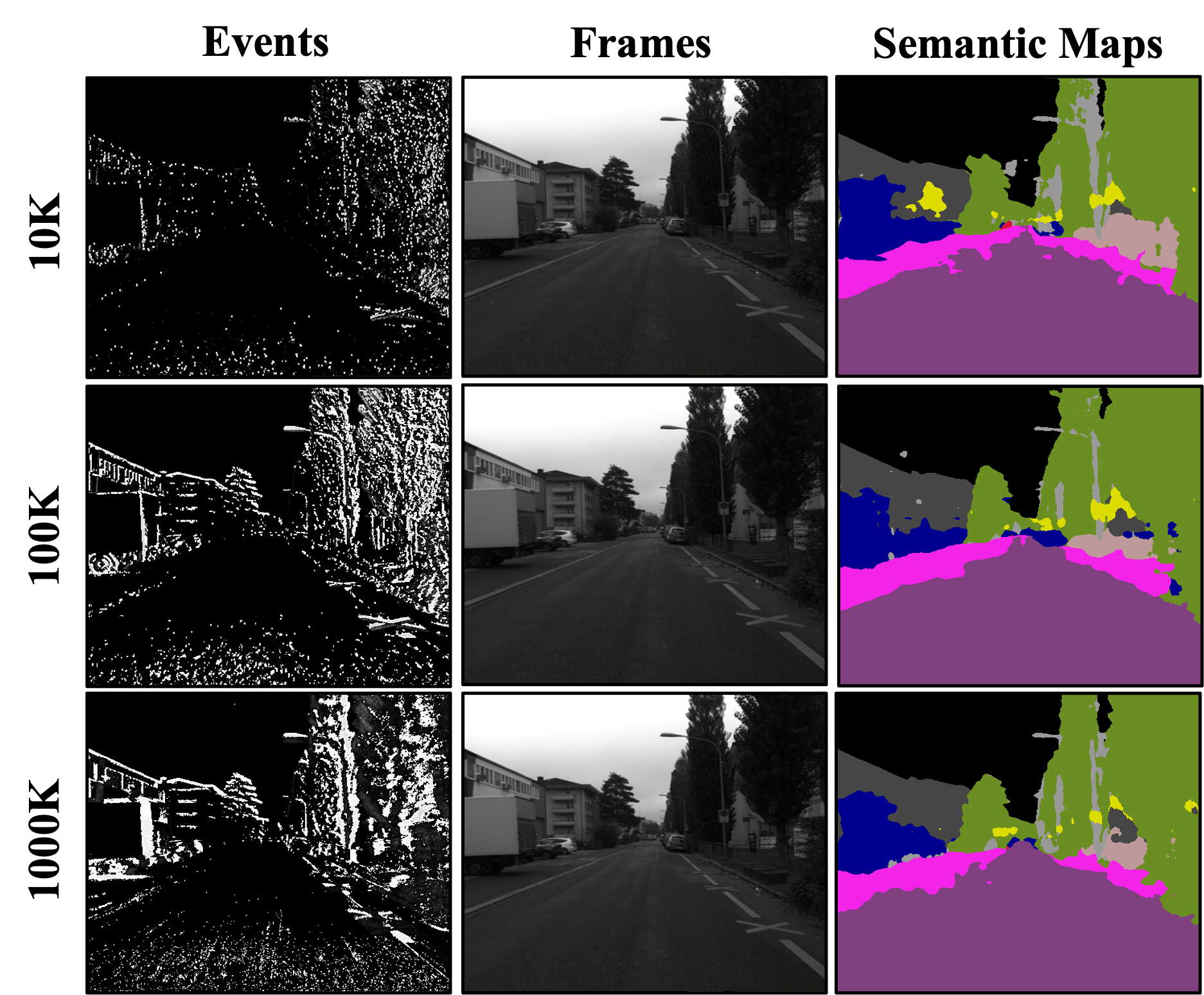}
    \caption{\textbf{Impact of event window density on predictions.} Top: $10K$; Middle: $100K$; Bottom: $1000K$. Moderate event bin density $100K$ offers better performance compared to its counterparts.}
    \vspace{-8mm}
    \label{fig:ablations}
  \hfill
\end{figure}
\subsection{Computing Approximate Inference Cost}
\label{sec:suppl-4}
This section validates the energy-efficiency of our method in terms of compute intensity per inference cycle. Spiking Neural Networks (SNNs), often referred to as the third generation of neural networks, are well known for being highly energy-efficient compared to traditional Artificial Neural Networks (ANNs). For estimating the inference cost per cycle for different architectures, we try to highlight how computations in SNNs and ANNs primarily differ from each other. SNNs offer highly sparse, asynchronous event-driven ACcumulate (AC) operations over time. Hence, the synaptic computes are performed only when an input spike arrives. In contrast, ANNs perform expensive Multiply-and-ACcumulate (MAC) operations for computing dense Matrix-Vector Multiplication (MVM) functions, irrespective of the sparsity of inputs. We use the findings in~\cite{mac} to specify that a MAC operation requires a total of $E_{MAC} =4.6pJ$ of energy while an AC operation requires only $E_{AC} =0.9pJ$ for a 32-bit floating-point computation in 45nm CMOS technology. This makes an AC operation $5.1\times$ more energy-efficient than a corresponding MAC operation. Note that comparisons on a different technology node would also generate similar energy requirement trends between SNNs and ANNs. Coupled with the number of floating point operations (FLOPs) performed by the network for a single inference, we benchmark compute energy cost of existing approaches following the estimation method used in~\cite{lee2020spike,kim2022beyond,lee2022fusion,merolla2014million, rueckauer2017conversion} on a 45nm CMOS process node.

It is worth mentioning that we neglect energy consumed by the memory or any peripheral circuitry and only consider the compute cost for MAC/ AC operations. First, we calculate the total number of synaptic operations performed in each layer. For SNNs, the number of FLOPs at a layer is obtained by multiplying the mean spiking rate at each timestep for that layer, the number of synaptic connections and the number of operating timesteps. The small input spiking activities obtained in different SNN layers are mainly because of the fact that event camera outputs are highly sparse in nature and the spiking neurons generate progressively sparser outputs as the network depth
increases. This sparse firing rate is essential for exploiting efficient event-based computations in the SNN layers. In contrast, ANNs execute dense matrix-vector multiplication operations without considering the sparsity of inputs. In other words, ANNs simply feed-forward the inputs at once, with a fixed total number of operations. This leads to high energy requirements (compared to SNNs) since operations are executed for both zero and non-zero input values, leading to unnecessary compute~\cite{lee2020spike}.

Given a number of neurons $M$, a number of synaptic connections $C$, and a mean firing activity $F$, the number of FLOPs at each timestep for a layer $l$ is calculated as $M_l \times C_l \times F_l$. In the case of ANNs, we have a $mean\_spiking\_rate = 1$ and $number\_of\_timesteps = 1$ for each layer. Hence, the total compute energy cost per inference cycle can be formalized as follows:
\begin{center}
\vspace{-6mm}
\[FLOPs_{ANN} = \sum_l M_l \times C_l\]\\ \vspace{-6mm}
\[FLOPs_{SNN}=N \sum_l M_l \times C_l \times F_l\]\\ \vspace{-6mm}
\[E_{Total} = FLOPs_{ANN} \times E_{MAC} + FLOPs_{SNN} \times E_{AC}\]
\end{center}

where $N$ is the number of timesteps and $E_{Total}$ denotes the total compute cost for a single inference. We utilize the above-described formulation to estimate the total compute energy required by different networks during inference. Note that we only consider convolution operations to compute the number of FLOPs, and neglect energy consumed by Batch-norm layers, or activations after each convolution layer. 
The results in Tab. 1, Tab. 2 and Tab. 3 in the main manuscript suggest that our method achieves competitive
performance and the least compute energy requirement compared to the `heavier' current state-of-the-art approaches~\cite{wang2021dual, alonso2019ev, gehrig2020video, wang2021evdistill, rebecq2019high}. Specifically, we reduce parameter count ($\sim$ 33$\times$ lower) and inference cost ($\sim$ 20$\times$ lower) compared to prior art. This is mainly attributed to our efficient hybrid SNN-ANN temporal-spatial feature learning method which enables us to use a lightweight architecture for edge-compute without compromising on semantic performance. Additionally, the SNN-encoder pathway in our network contributes negligibly to the total compute energy cost, reducing our energy requirements even further.

\subsection{Architecture Details}
\label{sec:suppl-5}
We provide details on the decoupled sampling rates $r_h \times r_w$ of the dilated convolution blocks used in our MMix module as follows: $1 \times 6$, ($6 \times 21$, $18 \times 15$, $1 \times 1$) and $6 \times 3$. The rates correspond to dilated conv. blocks appearing from left to right in Fig. 4 from the main paper. For blocks appearing in the same vertical alignment, rates are stated in top-to-bottom order inside single brackets.

\subsection{Additional visualisations}
\label{sec:suppl-6}
Following the discussion in the main paper, we provide visualisations for segmentation results using only one modality (events or frames) versus multi-modal settings in~\cref{fig:fused} and show the impact of varying event window density on predictions in~\cref{fig:ablations}.
{\small
\bibliographystyle{ieeetr}
\bibliography{egbib}

\begin{thebibliography}{10}

\bibitem{borst2010fly}
A.~Borst, J.~Haag, and D.~F. Reiff, ``Fly motion vision,'' {\em Annual review
  of neuroscience}, vol.~33, pp.~49--70, 2010.

\bibitem{sze2017efficient}
V.~Sze, Y.-H. Chen, T.-J. Yang, and J.~S. Emer, ``Efficient processing of deep
  neural networks: A tutorial and survey,'' {\em Proceedings of the IEEE},
  vol.~105, no.~12, pp.~2295--2329, 2017.

\bibitem{honeybee1996}
M.~Srinivasan, S.~Zhang, M.~Lehrer, and T.~Collett, ``Honeybee navigation en
  route to the goal: visual flight control and odometry,'' {\em Journal of
  Experimental Biology}, vol.~199, no.~1, pp.~237--244, 1996.

\bibitem{baird2013universal}
E.~Baird, N.~Boeddeker, M.~R. Ibbotson, and M.~V. Srinivasan, ``A universal
  strategy for visually guided landing,'' {\em Proceedings of the National
  Academy of Sciences}, vol.~110, no.~46, pp.~18686--18691, 2013.

\bibitem{lichtsteiner}
P.~Lichtsteiner, C.~Posch, and T.~Delbruck, ``A 128$\times$ 128 120 db 15
  $\mu$s latency asynchronous temporal contrast vision sensor,'' {\em IEEE
  Journal of Solid-State Circuits}, vol.~43, no.~2, pp.~566--576, 2008.

\bibitem{brandli2014240}
C.~Brandli, R.~Berner, M.~Yang, S.-C. Liu, and T.~Delbruck, ``A 240$\times$ 180
  130 db 3 $\mu$s latency global shutter spatiotemporal vision sensor,'' {\em
  IEEE Journal of Solid-State Circuits}, vol.~49, no.~10, pp.~2333--2341, 2014.

\bibitem{delbruck2010activity}
T.~Delbr{\"u}ck, B.~Linares-Barranco, E.~Culurciello, and C.~Posch,
  ``Activity-driven, event-based vision sensors,'' in {\em Proceedings of 2010
  IEEE International Symposium on Circuits and Systems}, pp.~2426--2429, IEEE,
  2010.

\bibitem{lee2020spike}
C.~Lee, A.~K. Kosta, A.~Z. Zhu, K.~Chaney, K.~Daniilidis, and K.~Roy,
  ``Spike-flownet: event-based optical flow estimation with energy-efficient
  hybrid neural networks,'' in {\em European Conference on Computer Vision},
  pp.~366--382, Springer, 2020.

\bibitem{tao2020hierarchical}
A.~Tao, K.~Sapra, and B.~Catanzaro, ``Hierarchical multi-scale attention for
  semantic segmentation,'' {\em arXiv preprint arXiv:2005.10821}, 2020.

\bibitem{yuan2019segmentation}
Y.~Yuan, X.~Chen, X.~Chen, and J.~Wang, ``Segmentation transformer:
  Object-contextual representations for semantic segmentation,'' {\em arXiv
  preprint arXiv:1909.11065}, 2019.

\bibitem{yuan2021ocnet}
Y.~Yuan, L.~Huang, J.~Guo, C.~Zhang, X.~Chen, and J.~Wang, ``Ocnet: Object
  context for semantic segmentation,'' {\em International Journal of Computer
  Vision}, vol.~129, no.~8, pp.~2375--2398, 2021.

\bibitem{zhao2017pyramid}
H.~Zhao, J.~Shi, X.~Qi, X.~Wang, and J.~Jia, ``Pyramid scene parsing network,''
  in {\em Proceedings of the IEEE conference on computer vision and pattern
  recognition}, pp.~2881--2890, 2017.

\bibitem{chen2018encoder}
L.-C. Chen, Y.~Zhu, G.~Papandreou, F.~Schroff, and H.~Adam, ``Encoder-decoder
  with atrous separable convolution for semantic image segmentation,'' in {\em
  Proceedings of the European conference on computer vision (ECCV)},
  pp.~801--818, 2018.

\bibitem{gallego2020event}
G.~Gallego, T.~Delbr{\"u}ck, G.~Orchard, C.~Bartolozzi, B.~Taba, A.~Censi,
  S.~Leutenegger, A.~J. Davison, J.~Conradt, K.~Daniilidis, {\em et~al.},
  ``Event-based vision: A survey,'' {\em IEEE transactions on pattern analysis
  and machine intelligence}, vol.~44, no.~1, pp.~154--180, 2020.

\bibitem{lagorce2016hots}
X.~Lagorce, G.~Orchard, F.~Galluppi, B.~E. Shi, and R.~B. Benosman, ``Hots: a
  hierarchy of event-based time-surfaces for pattern recognition,'' {\em IEEE
  transactions on pattern analysis and machine intelligence}, vol.~39, no.~7,
  pp.~1346--1359, 2016.

\bibitem{maqueda2018event}
A.~I. Maqueda, A.~Loquercio, G.~Gallego, N.~Garc{\'\i}a, and D.~Scaramuzza,
  ``Event-based vision meets deep learning on steering prediction for
  self-driving cars,'' in {\em Proceedings of the IEEE conference on computer
  vision and pattern recognition}, pp.~5419--5427, 2018.

\bibitem{rebecq2017real}
H.~Rebecq, T.~Horstschaefer, and D.~Scaramuzza, ``Real-time visual-inertial
  odometry for event cameras using keyframe-based nonlinear optimization,''
  2017.

\bibitem{sun2022ess}
Z.~Sun, N.~Messikommer, D.~Gehrig, and D.~Scaramuzza, ``Ess: Learning
  event-based semantic segmentation from still images,'' {\em arXiv preprint
  arXiv:2203.10016}, 2022.

\bibitem{rebecq2019high}
H.~Rebecq, R.~Ranftl, V.~Koltun, and D.~Scaramuzza, ``High speed and high
  dynamic range video with an event camera,'' {\em IEEE transactions on pattern
  analysis and machine intelligence}, vol.~43, no.~6, pp.~1964--1980, 2019.

\bibitem{roy2019towards}
K.~Roy, A.~Jaiswal, and P.~Panda, ``Towards spike-based machine intelligence
  with neuromorphic computing,'' {\em Nature}, vol.~575, no.~7784,
  pp.~607--617, 2019.

\bibitem{wang2022hierarchical}
S.~Wang, T.~H. Cheng, and M.~H. Lim, ``A hierarchical taxonomic survey of
  spiking neural networks,'' {\em Memetic Computing}, vol.~14, no.~3,
  pp.~335--354, 2022.

\bibitem{ponghiran2022spiking}
W.~Ponghiran and K.~Roy, ``Spiking neural networks with improved inherent
  recurrence dynamics for sequential learning,'' in {\em Proceedings of the
  AAAI Conference on Artificial Intelligence}, vol.~36, pp.~8001--8008, 2022.

\bibitem{davies2018loihi}
M.~Davies, N.~Srinivasa, T.-H. Lin, G.~Chinya, Y.~Cao, S.~H. Choday, G.~Dimou,
  P.~Joshi, N.~Imam, S.~Jain, {\em et~al.}, ``Loihi: A neuromorphic manycore
  processor with on-chip learning,'' {\em Ieee Micro}, vol.~38, no.~1,
  pp.~82--99, 2018.

\bibitem{akopyan2015truenorth}
F.~Akopyan, J.~Sawada, A.~Cassidy, R.~Alvarez-Icaza, J.~Arthur, P.~Merolla,
  N.~Imam, Y.~Nakamura, P.~Datta, G.-J. Nam, {\em et~al.}, ``Truenorth: Design
  and tool flow of a 65 mw 1 million neuron programmable neurosynaptic chip,''
  {\em IEEE transactions on computer-aided design of integrated circuits and
  systems}, vol.~34, no.~10, pp.~1537--1557, 2015.

\bibitem{furber2014spinnaker}
S.~B. Furber, F.~Galluppi, S.~Temple, and L.~A. Plana, ``The spinnaker
  project,'' {\em Proceedings of the IEEE}, vol.~102, no.~5, pp.~652--665,
  2014.

\bibitem{binas2017ddd17}
J.~Binas, D.~Neil, S.-C. Liu, and T.~Delbruck, ``Ddd17: End-to-end davis
  driving dataset,'' {\em arXiv preprint arXiv:1711.01458}, 2017.

\bibitem{zhu2018multivehicle}
A.~Z. Zhu, D.~Thakur, T.~{\"O}zaslan, B.~Pfrommer, V.~Kumar, and K.~Daniilidis,
  ``The multivehicle stereo event camera dataset: An event camera dataset for
  3d perception,'' {\em IEEE Robotics and Automation Letters}, vol.~3, no.~3,
  pp.~2032--2039, 2018.

\bibitem{alonso2019ev}
I.~Alonso and A.~C. Murillo, ``Ev-segnet: Semantic segmentation for event-based
  cameras,'' in {\em Proceedings of the IEEE/CVF Conference on Computer Vision
  and Pattern Recognition Workshops}, pp.~0--0, 2019.

\bibitem{chollet2017xception}
F.~Chollet, ``Xception: Deep learning with depthwise separable convolutions,''
  in {\em Proceedings of the IEEE conference on computer vision and pattern
  recognition}, pp.~1251--1258, 2017.

\bibitem{gehrig2020video}
D.~Gehrig, M.~Gehrig, J.~Hidalgo-Carri{\'o}, and D.~Scaramuzza, ``Video to
  events: Recycling video datasets for event cameras,'' in {\em Proceedings of
  the IEEE/CVF Conference on Computer Vision and Pattern Recognition},
  pp.~3586--3595, 2020.

\bibitem{wang2021evdistill}
L.~Wang, Y.~Chae, S.-H. Yoon, T.-K. Kim, and K.-J. Yoon, ``Evdistill:
  Asynchronous events to end-task learning via bidirectional
  reconstruction-guided cross-modal knowledge distillation,'' in {\em
  Proceedings of the IEEE/CVF Conference on Computer Vision and Pattern
  Recognition}, pp.~608--619, 2021.

\bibitem{cordts2016cityscapes}
M.~Cordts, M.~Omran, S.~Ramos, T.~Rehfeld, M.~Enzweiler, R.~Benenson,
  U.~Franke, S.~Roth, and B.~Schiele, ``The cityscapes dataset for semantic
  urban scene understanding,'' in {\em Proceedings of the IEEE conference on
  computer vision and pattern recognition}, pp.~3213--3223, 2016.

\bibitem{wang2021dual}
L.~Wang, Y.~Chae, and K.-J. Yoon, ``Dual transfer learning for event-based
  end-task prediction via pluggable event to image translation,'' in {\em
  Proceedings of the IEEE/CVF International Conference on Computer Vision},
  pp.~2135--2145, 2021.

\bibitem{messikommer2022bridging}
N.~Messikommer, D.~Gehrig, M.~Gehrig, and D.~Scaramuzza, ``Bridging the gap
  between events and frames through unsupervised domain adaptation,'' {\em IEEE
  Robotics and Automation Letters}, vol.~7, no.~2, pp.~3515--3522, 2022.

\bibitem{thanh2020catastrophic}
H.~Thanh-Tung and T.~Tran, ``Catastrophic forgetting and mode collapse in
  gans,'' in {\em 2020 international joint conference on neural networks
  (ijcnn)}, pp.~1--10, IEEE, 2020.

\bibitem{Baltrušaitismultimodal}
T.~Baltrušaitis, C.~Ahuja, and L.-P. Morency, ``Multimodal machine learning: A
  survey and taxonomy,'' {\em IEEE Transactions on Pattern Analysis and Machine
  Intelligence}, vol.~41, no.~2, pp.~423--443, 2019.

\bibitem{neil2016phased}
D.~Neil, M.~Pfeiffer, and S.-C. Liu, ``Phased lstm: Accelerating recurrent
  network training for long or event-based sequences,'' {\em Advances in neural
  information processing systems}, vol.~29, 2016.

\bibitem{oviatt2018handbook}
S.~Oviatt, B.~Schuller, P.~R. Cohen, D.~Sonntag, G.~Potamianos, and
  A.~Kr{\"u}ger, {\em The Handbook of Multimodal-Multisensor Interfaces: Signal
  Processing, Architectures, and Detection of Emotion and Cognition-Volume 2}.
\newblock Association for Computing Machinery and Morgan \& Claypool, 2018.

\bibitem{kim2022beyond}
Y.~Kim, J.~Chough, and P.~Panda, ``Beyond classification: directly training
  spiking neural networks for semantic segmentation,'' {\em Neuromorphic
  Computing and Engineering}, 2022.

\bibitem{chen2017deeplab}
L.-C. Chen, G.~Papandreou, I.~Kokkinos, K.~Murphy, and A.~L. Yuille, ``Deeplab:
  Semantic image segmentation with deep convolutional nets, atrous convolution,
  and fully connected crfs,'' {\em IEEE transactions on pattern analysis and
  machine intelligence}, vol.~40, no.~4, pp.~834--848, 2017.

\bibitem{chen2014semantic}
L.-C. Chen, G.~Papandreou, I.~Kokkinos, K.~Murphy, and A.~L. Yuille, ``Semantic
  image segmentation with deep convolutional nets and fully connected crfs,''
  {\em arXiv preprint arXiv:1412.7062}, 2014.

\bibitem{pei2019towards}
J.~Pei, L.~Deng, S.~Song, M.~Zhao, Y.~Zhang, S.~Wu, G.~Wang, Z.~Zou, Z.~Wu,
  W.~He, {\em et~al.}, ``Towards artificial general intelligence with hybrid
  tianjic chip architecture,'' {\em Nature}, vol.~572, no.~7767, pp.~106--111,
  2019.

\bibitem{jaderberg2015}
M.~Jaderberg, K.~Simonyan, A.~Zisserman, and k.~kavukcuoglu, ``Spatial
  transformer networks,'' in {\em Advances in Neural Information Processing
  Systems} (C.~Cortes, N.~Lawrence, D.~Lee, M.~Sugiyama, and R.~Garnett, eds.),
  vol.~28, Curran Associates, Inc., 2015.

\bibitem{wang2019event}
L.~Wang, Y.-S. Ho, K.-J. Yoon, {\em et~al.}, ``Event-based high dynamic range
  image and very high frame rate video generation using conditional generative
  adversarial networks,'' in {\em Proceedings of the IEEE/CVF Conference on
  Computer Vision and Pattern Recognition}, pp.~10081--10090, 2019.

\bibitem{baldwin2022time}
R.~Baldwin, R.~Liu, M.~M. Almatrafi, V.~K. Asari, and K.~Hirakawa,
  ``Time-ordered recent event (tore) volumes for event cameras,'' {\em IEEE
  Transactions on Pattern Analysis and Machine Intelligence}, 2022.

\bibitem{barchid2022bina}
S.~Barchid, J.~Mennesson, and C.~Dj{\'e}raba, ``Bina-rep event frames: A simple
  and effective representation for event-based cameras,'' in {\em 2022 IEEE
  International Conference on Image Processing (ICIP)}, pp.~3998--4002, IEEE,
  2022.

\bibitem{abbott1999lapicque}
L.~F. Abbott, ``Lapicque’s introduction of the integrate-and-fire model
  neuron (1907),'' {\em Brain research bulletin}, vol.~50, no.~5-6,
  pp.~303--304, 1999.

\bibitem{dayan2001theoretical}
P.~Dayan, L.~F. Abbott, {\em et~al.}, ``Theoretical neuroscience (vol. 806),''
  2001.

\bibitem{weisstein2002heaviside}
E.~W. Weisstein, ``Heaviside step function,'' {\em https://mathworld. wolfram.
  com/}, 2002.

\bibitem{ledinauskas2020training}
E.~Ledinauskas, J.~Ruseckas, A.~Jur{\v{s}}{\.e}nas, and G.~Bura{\v{c}}as,
  ``Training deep spiking neural networks,'' {\em arXiv preprint
  arXiv:2006.04436}, 2020.

\bibitem{han2020rmp}
B.~Han, G.~Srinivasan, and K.~Roy, ``Rmp-snn: Residual membrane potential
  neuron for enabling deeper high-accuracy and low-latency spiking neural
  network,'' in {\em Proceedings of the IEEE/CVF conference on computer vision
  and pattern recognition}, pp.~13558--13567, 2020.

\bibitem{fontaine2014spike}
B.~Fontaine, J.~L. Pe{\~n}a, and R.~Brette, ``Spike-threshold adaptation
  predicted by membrane potential dynamics in vivo,'' {\em PLoS computational
  biology}, vol.~10, no.~4, p.~e1003560, 2014.

\bibitem{pozo2010unraveling}
K.~Pozo and Y.~Goda, ``Unraveling mechanisms of homeostatic synaptic
  plasticity,'' {\em Neuron}, vol.~66, no.~3, pp.~337--351, 2010.

\bibitem{rathi2021diet}
N.~Rathi and K.~Roy, ``Diet-snn: A low-latency spiking neural network with
  direct input encoding and leakage and threshold optimization,'' {\em IEEE
  Transactions on Neural Networks and Learning Systems}, 2021.

\bibitem{mac}
M.~Horowitz, ``1.1 computing's energy problem (and what we can do about it),''
  in {\em 2014 IEEE International Solid-State Circuits Conference Digest of
  Technical Papers (ISSCC)}, pp.~10--14, 2014.

\bibitem{ioffe2015batch}
S.~Ioffe and C.~Szegedy, ``Batch normalization: Accelerating deep network
  training by reducing internal covariate shift,'' in {\em International
  conference on machine learning}, pp.~448--456, PMLR, 2015.

\bibitem{xu2015empirical}
B.~Xu, N.~Wang, T.~Chen, and M.~Li, ``Empirical evaluation of rectified
  activations in convolutional network,'' {\em arXiv preprint
  arXiv:1505.00853}, 2015.

\bibitem{kingma2015adam}
D.~P. Kingma and J.~Ba, ``Adam: {A} method for stochastic optimization,'' in
  {\em 3rd International Conference on Learning Representations, {ICLR} 2015,
  San Diego, CA, USA, May 7-9, 2015, Conference Track Proceedings} (Y.~Bengio
  and Y.~LeCun, eds.), 2015.

\bibitem{neftci2019surrogate}
E.~O. Neftci, H.~Mostafa, and F.~Zenke, ``Surrogate gradient learning in
  spiking neural networks: Bringing the power of gradient-based optimization to
  spiking neural networks,'' {\em IEEE Signal Processing Magazine}, vol.~36,
  no.~6, pp.~51--63, 2019.

\bibitem{lee2016training}
J.~H. Lee, T.~Delbruck, and M.~Pfeiffer, ``Training deep spiking neural
  networks using backpropagation,'' {\em Frontiers in neuroscience}, vol.~10,
  p.~508, 2016.

\bibitem{deng2020rethinking}
L.~Deng, Y.~Wu, X.~Hu, L.~Liang, Y.~Ding, G.~Li, G.~Zhao, P.~Li, and Y.~Xie,
  ``Rethinking the performance comparison between snns and anns,'' {\em Neural
  networks}, vol.~121, pp.~294--307, 2020.

\bibitem{gehrig2021dsec}
M.~Gehrig, W.~Aarents, D.~Gehrig, and D.~Scaramuzza, ``Dsec: A stereo event
  camera dataset for driving scenarios,'' {\em IEEE Robotics and Automation
  Letters}, vol.~6, no.~3, pp.~4947--4954, 2021.

\bibitem{shi2015convolutional}
X.~Shi, Z.~Chen, H.~Wang, D.-Y. Yeung, W.-K. Wong, and W.-c. Woo,
  ``Convolutional lstm network: A machine learning approach for precipitation
  nowcasting,'' {\em Advances in neural information processing systems},
  vol.~28, 2015.

\bibitem{li2022asynchronous}
J.~Li, J.~Li, L.~Zhu, X.~Xiang, T.~Huang, and Y.~Tian, ``Asynchronous
  spatio-temporal memory network for continuous event-based object detection,''
  {\em IEEE Transactions on Image Processing}, vol.~31, pp.~2975--2987, 2022.

\bibitem{lee2022fusion}
C.~Lee, A.~K. Kosta, and K.~Roy, ``Fusion-flownet: Energy-efficient optical
  flow estimation using sensor fusion and deep fused spiking-analog network
  architectures,'' in {\em 2022 International Conference on Robotics and
  Automation (ICRA)}, pp.~6504--6510, IEEE, 2022.

\bibitem{merolla2014million}
P.~A. Merolla, J.~V. Arthur, R.~Alvarez-Icaza, A.~S. Cassidy, J.~Sawada,
  F.~Akopyan, B.~L. Jackson, N.~Imam, C.~Guo, Y.~Nakamura, {\em et~al.}, ``A
  million spiking-neuron integrated circuit with a scalable communication
  network and interface,'' {\em Science}, vol.~345, no.~6197, pp.~668--673,
  2014.

\bibitem{rueckauer2017conversion}
B.~Rueckauer, I.-A. Lungu, Y.~Hu, M.~Pfeiffer, and S.-C. Liu, ``Conversion of
  continuous-valued deep networks to efficient event-driven networks for image
  classification,'' {\em Frontiers in neuroscience}, vol.~11, p.~682, 2017.

\end{thebibliography}
}
\end{document}